%% file: main.tex
\documentclass[letterpaper]{article} 
\usepackage{aaai23}
\usepackage{times}  
\usepackage{helvet}  
\usepackage{courier}  
\usepackage[hyphens]{url}  
\usepackage{graphicx} 
\usepackage{natbib}  
\usepackage{caption} 
\title{Predicate Invention for Bilevel Planning}
\author{
    Tom Silver\equalcontrib$^1$,
    Rohan Chitnis\equalcontrib$^2$,
    Nishanth Kumar$^1$,
    Willie McClinton$^1$,\\
    Tom\'as Lozano-P\'erez$^1$,
    Leslie Pack Kaelbling$^1$,
    Joshua B. Tenenbaum$^1$
}
\affiliations{
    $^1$MIT Computer Science and Artificial 
  Intelligence Laboratory\quad$^2$Meta AI\\

    \{tslvr, njk, wbm3, tlp, lpk, jbt\}@mit.edu, ronuchit@meta.com
}

\input{preamble}

\input{variables}

\begin{document}

\maketitle

\input{abstract}

\section{Introduction}
\label{sec:intro}

\input{introduction}

\section{Problem Setting}
\label{sec:problem}

\input{problem}

\section{Predicates, Operators, and Samplers}
\label{sec:representation}

\input{representation}

\section{Bilevel Planning}
\label{sec:planning}

\input{planning}

\section{Learning from Demonstrations}
\label{sec:learning}

\input{learning}

\section{Experiments}
\label{sec:experiments}

\input{experiments}

\section{Related Work}
\label{sec:rw}

\input{related_work}

\section{Conclusion and Future Work}
\label{sec:conclusion}

\input{conclusion}

\clearpage
\section*{Acknowledgements}
\input{acknowledgements}

\bibliography{references}

\appendix

\clearpage
\section{Appendix}
\label{sec:Appendix}

\input{appendix}

\end{document}

%% file: preamble.tex
\usepackage{wrapfig}

\usepackage{color}
\usepackage{amssymb}
\usepackage{fancyvrb}
\usepackage{pmboxdraw}
\usepackage{amsmath}
\usepackage{amsthm}
\usepackage{tikz}
\usepackage[hang,flushmargin]{footmisc}
\usepackage{chngcntr}
\usepackage[makeroom]{cancel}
\usetikzlibrary{arrows,decorations.markings}

\usepackage{booktabs}
\usepackage{graphicx}
\usepackage{mathtools}
\usepackage{bbm}
\usepackage{relsize}

\usepackage{caption}
\usepackage{subcaption}

\usepackage{algorithm2e}
\let\oldnl\nl
\newcommand{\nonl}{\renewcommand{\nl}{\let\nl\oldnl}}
\makeatletter
\newcommand{\removelatexerror}{\let\@latex@error\@gobble}
\makeatother

\usepackage{enumitem}

\newenvironment{tightlist}%
{\begin{list}{$\bullet$}{%
    \setlength{\topsep}{0in}
    \setlength{\partopsep}{0in}
    \setlength{\itemsep}{0in}
    \setlength{\parsep}{0in}
    \setlength{\leftmargin}{1.5em}
    \setlength{\rightmargin}{0in}
}
}%
{\end{list}
}

\newcommand{\secref}[1]{Section \ref{#1}}

\newcommand{\figref}[1]{Figure~\ref{#1}}
\newcommand{\algref}[1]{Algorithm~\ref{#1}}
\newcommand{\algrefshort}[1]{Alg.~\ref{#1}}
\newcommand{\tabref}[1]{Table~\ref{#1}}

\newcommand{\defref}[1]{Definition~\ref{#1}}
\newcommand{\appref}[1]{Appendix~\ref{#1}}

\definecolor{green}{RGB}{0, 150, 0}

\newtheorem{defn}{Definition}

\counterwithin*{thm}{section}

\usepackage{mdframed}
\newmdtheoremenv[nobreak=true]{def2}{Definition}

\def\thickhline{%
  \noalign{\ifnum0=`}\fi\hrule \@height \thickarrayrulewidth \futurelet
  \reserved@a\@xthickhline}
\def\@xthickhline{\ifx\reserved@a\thickhline
              \vskip\doublerulesep
              \vskip-\thickarrayrulewidth
             \fi
      \ifnum0=`{\fi}}

\newlength{\thickarrayrulewidth}
\DeclareUnicodeCharacter{225C}{$\triangleq$}
\DeclareUnicodeCharacter{2200}{$\forall$}

%% file: variables.tex
\renewcommand{\O}{\mathcal{O}}

\renewcommand{\S}{\mathcal{S}} 
\newcommand{\A}{\mathcal{A}}
\newcommand{\T}{\mathcal{T}}

\newcommand{\X}{\mathcal{X}}

\newcommand{\C}{\mathcal{C}}

\newcommand{\D}{\mathcal{D}}

\DeclareMathOperator*{\argmin}{argmin}

\DeclarePairedDelimiterX{\infdivx}[2]{(}{)}{%
  #1\;\delimsize\|\;#2%
}

\newcommand{\types}{\Lambda}
\newcommand{\type}{\lambda}

\newcommand{\params}{\textsc{Par}}
\newcommand{\preconditions}{\textsc{Pre}}
\newcommand{\addeffects}{\textsc{Eff}^{+}}
\newcommand{\deleteeffects}{\textsc{Eff}^{-}}
\newcommand{\controllerspec}{\textsc{Con}}
\newcommand{\substitution}{\delta}
\newcommand{\operator}{\omega}
\newcommand{\operators}{\Omega}
\newcommand{\ground}{\underline}
\newcommand{\sampler}{\sigma}
\newcommand{\samplers}{\Sigma}

%% file: abstract.tex
\begin{abstract}
Efficient planning in continuous state and action spaces is fundamentally hard, even when the transition model is deterministic and known.
One way to alleviate this challenge is to perform bilevel planning with abstractions, where a high-level search for abstract plans is used to guide planning in the original transition space.
Previous work has shown that when state abstractions in the form of symbolic predicates are hand-designed, operators and samplers for bilevel planning can be learned from demonstrations.
In this work, we propose an algorithm for learning predicates from demonstrations, eliminating the need for manually specified state abstractions.
Our key idea is to learn predicates by optimizing a surrogate objective that is tractable but faithful to our real efficient-planning objective.
We use this surrogate objective in a hill-climbing search over predicate sets drawn from a grammar.
Experimentally, we show across four robotic planning environments that our learned abstractions are able to quickly solve held-out tasks, outperforming six baselines.
\end{abstract}

%% file: introduction.tex
Hierarchical planning is a powerful approach for decision-making in environments with continuous states, continuous actions, and long horizons.
A crucial bottleneck in scaling hierarchical planning is the reliance on human engineers to manually program domain-specific abstractions.
For example, in bilevel sample-based task and motion planning~\cite{srivastava2014combined,garrett2021integrated}, an engineer must design (1) symbolic predicates; (2) symbolic operators; and (3) samplers that propose different refinements of the symbolic operators into continuous actions.
However, recent work has shown that when predicates are \emph{given}, operators and samplers can be learned from a modest number (50--200) of demonstrations~\cite{silver2021learning,chitnis2021learning}.
Our objective in this work is to \emph{learn} predicates that can then be used to learn operators and samplers.

\begin{figure*}[t]
  \centering
    \noindent
    \includegraphics[width=\textwidth]{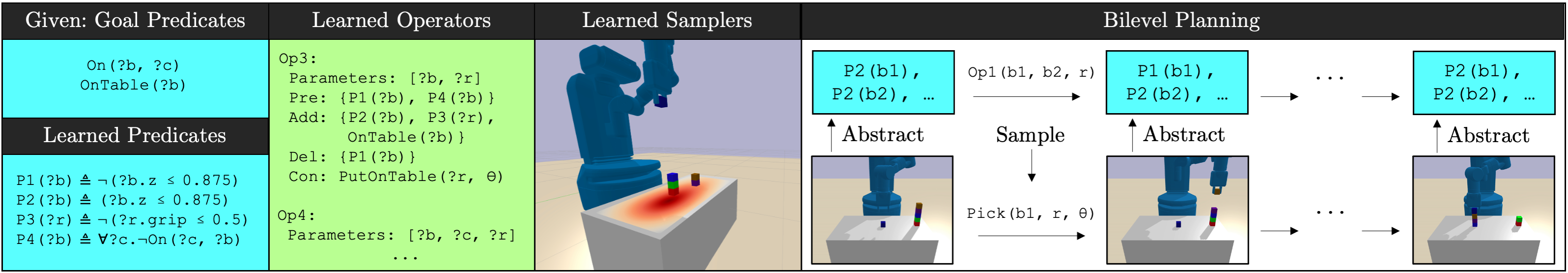}
    \caption{\small{\textbf{Overview of our framework.} Given a small set of goal predicates (first panel, top), we use demonstration data to learn new predicates (first panel, bottom). In this Blocks example, the learned predicates \texttt{P1} -- \texttt{P4} intuitively represent \texttt{Holding}, \texttt{NotHolding}, \texttt{HandEmpty}, and \texttt{NothingAbove} respectively. Collectively, the predicates define a state abstraction that maps continuous states $x$ in the environment to abstract states $s$. Object types are omitted for clarity. After predicate invention, we learn abstractions of the continuous action space and transition model via planning operators (second panel). For each operator, we learn a sampler (third panel), a neural network that maps continuous object features in a given state to continuous action parameters for controllers which can be executed in the environment. In this example, the sampler proposes different placements on the table for the held block.
    With these learned representations, we perform bilevel planning (fourth panel), with search in the abstract spaces guiding planning in the continuous spaces.}}
  \label{fig:teaser}
\end{figure*}

Predicates in bilevel planning represent a discrete state abstraction of the underlying continuous state space~\cite{li2006towards,abel2017near}.
For example, \texttt{On(block1, block2)} is an abstraction that discards the exact continuous poses of \texttt{block1} and \texttt{block2}.
State abstraction alone is useful for decision-making, but predicates go further: together with operators, predicates enable the use of highly-optimized domain-independent AI planners~\cite{helmert2006fast}.

We consider a problem setting where a small set of \emph{goal predicates} are available and sufficient for describing task goals, but practically insufficient for bilevel planning.
For example, in a block stacking domain~(\figref{fig:teaser}), we start with \texttt{On} and \texttt{OnTable}, but have no predicates for describing whether a block is currently held or graspable.
Our aim is to invent new predicates to enrich the state abstraction beyond what can be expressed with the goal predicates alone, leading to stronger reasoning at the abstract level.

What objective should we optimize to learn predicates for bilevel planning?
First, consider our real objective: we want a predicate set such that bilevel planning is fast and successful, in expectation over a task distribution, when we use those predicates to learn operators and samplers for planning.
Unfortunately, this real objective is far too expensive to use directly, since even a single evaluation requires neural network sampler training and bilevel planning.

In this work, we propose a novel surrogate objective that is deeply connected to our real bilevel-planning objective, but is tractable for predicate learning.
Our main insight is that demonstrations can be used to analytically approximate bilevel planning time.
To leverage this objective for predicate learning, we take inspiration from the program synthesis literature~\cite{menon2013machine,ellis2020dreamcoder}, and learn predicates via a hill-climbing search through a grammar, with the search guided by the objective.
After predicate learning, we use the predicates to learn operators and samplers.
All three components can then be used for efficient bilevel planning on new tasks.

In experiments across four robotic planning environments, we find predicates, operators, and samplers learned from 50--200 demonstrations enable efficient bilevel planning on held-out tasks that involve different numbers of objects, longer horizons, and larger goal expressions than seen in the demonstrations.
Furthermore, predicates learned with our proposed surrogate objective substantially outperform those learned with objectives inspired by previous work, which are based on prediction error~\cite{pasula2007learning,jetchev2013learning}, bisimulation~\cite{konidaris2018skills,curtis2021discovering}, and inverse planning~\cite{baker2009action,ramirez2010probabilistic,zhi2020online}.
We compare against several other baselines and ablations of our system to further validate our results.

%% file: problem.tex
We consider learning from demonstrations in deterministic planning problems.
These problems are goal-based and object-centric, with continuous states and hybrid discrete-continuous actions.
Formally, an \emph{environment} is a tuple $\langle \types, d, \C, f, \Psi_G \rangle$, and is associated with a distribution $\T$ over \emph{tasks}, where each task $T \in \T$ is a tuple $\langle \O, x_0, g \rangle$.

$\types$ is a finite set of object \emph{types}, and the map $d: \types \to \mathbb{N}$ defines the dimensionality of the real-valued feature vector for each type. Within a task, $\O$ is an \emph{object set}, where each object has a type drawn from $\types$; this $\O$ can (and typically will) vary between tasks.
$\O$ induces a state space $\X_\O$ (going forward, we simply write $\X$ when clear from context). A \emph{state} $x \in \X$ in a task is a mapping from each $o \in \O$ to a feature vector in $\mathbb{R}^{d(\text{type}(o))}$; $x_0$ is the initial state of the task.

$\C$ is a finite set of \emph{controllers}. A controller $C((\type_1, \dots, \type_v), \Theta) \in \C$ can have both discrete typed parameters $(\type_1, \ldots, \type_v)$ and a continuous real-valued vector of parameters $\Theta$. For instance, a controller \texttt{Pick} for picking up a block might have one discrete parameter of type \texttt{block} and a $\Theta$ that is a placeholder for a specific grasp pose. The controller set $\C$ and object set $\O$ induce an action space $\A_\O$ (going forward, we write $\A$ when clear). An \emph{action} $a \in \A$ in a task is a controller $C \in \C$ with both discrete and continuous arguments: $a = C((o_1, \ldots o_v), \theta)$, where the objects $(o_1, \ldots o_v)$ are drawn from the object set $\O$ and must have types matching the controller's discrete parameters $(\type_1, \ldots, \type_v)$. 
Transitions through states and actions are governed by $f: \X \times \A \to \X$, a known, deterministic transition model that is shared across tasks.

A \emph{predicate} $\psi$ is characterized by an ordered list of types $(\type_1, \dots, \type_m)$ and a lifted binary state classifier $c_\psi : \X \times \O^m \to \{\text{true}, \text{false}\}$, where $c_\psi(x, (o_1, \dots, o_m))$ is defined only when each object $o_i$ has type $\type_i$.
For instance, the predicate \mbox{\texttt{Holding}} may, given a state and two objects, robot and block, describe whether the block is held by the robot in this state.
A \emph{lifted atom} is a predicate with typed variables (e.g., \texttt{Holding(?robot, ?block)}).
A \emph{ground atom} $\ground{\psi}$ consists of a predicate $\psi$ and objects $(o_1, \dots, o_m)$, again with all $\text{type}(o_i) = \type_i$ (e.g., \texttt{Holding(robby, block7)}).
Note that a ground atom induces a binary state classifier $c_{\ground{\psi}} : \X \to \{\text{true}, \text{false}\}$, where $c_{\ground{\psi}}(x) \triangleq c_{\psi}(x, (o_1, \dots, o_m))$.

$\Psi_G$ is a small set of \emph{goal predicates} that we assume are given and sufficient for representing task goals, but insufficient practically as standalone state abstractions. Specifically, the goal $g$ of a task is a set of ground atoms over predicates in $\Psi_G$ and objects in $\O$.
A goal $g$ is said to \emph{hold} in a state $x$ if for all ground atoms $\ground{\psi} \in g$, the classifier $c_{\ground{\psi}}(x)$ returns true.
A solution to a task is a \emph{plan} $\pi = (a_1, \ldots, a_n)$, a sequence of actions $a \in \A$ such that successive application of the transition model $x_i = f(x_{i-1}, a_i)$ on each $a_i \in \pi$, starting from $x_0$, results in a final state $x_n$ where $g$ holds.

The agent is provided with a set of \emph{training tasks} from $\T$ and a set of demonstrations $\D$, with \emph{one demonstration per task}.
We assume action costs are unitary and demonstrations are near-optimal.
Each demonstration consists of a training task $\langle \O, x_0, g \rangle$ and a plan $\pi^*$ that solves the task.
Note that for each $\pi^*$, we can recover the associated state sequence starting at $x_0$, since $f$ is known and deterministic.
The agent's objective is to \emph{efficiently} solve held-out tasks from $\T$ using anything it chooses to learn from $\D$.

%% file: representation.tex
Since the agent has access to the transition model $f$, one approach for optimizing the objective described in \secref{sec:problem} is to forgo learning entirely, and solve any held-out task by running a planner over the state state $\X$ and action space $\A$.
However, searching for a solution directly in these large spaces is highly infeasible.
Instead, we propose to \emph{learn abstractions} using the provided demonstrations.
In this section, we will describe representations that allow for fast bilevel planning with abstractions (\secref{sec:planning}).
In \secref{sec:learning}, we then describe how to learn these abstractions.

\begin{wrapfigure}{r}{0.1\textwidth}
\vspace{1em}
\hspace{-1em}
\includegraphics[width=0.1\textwidth]{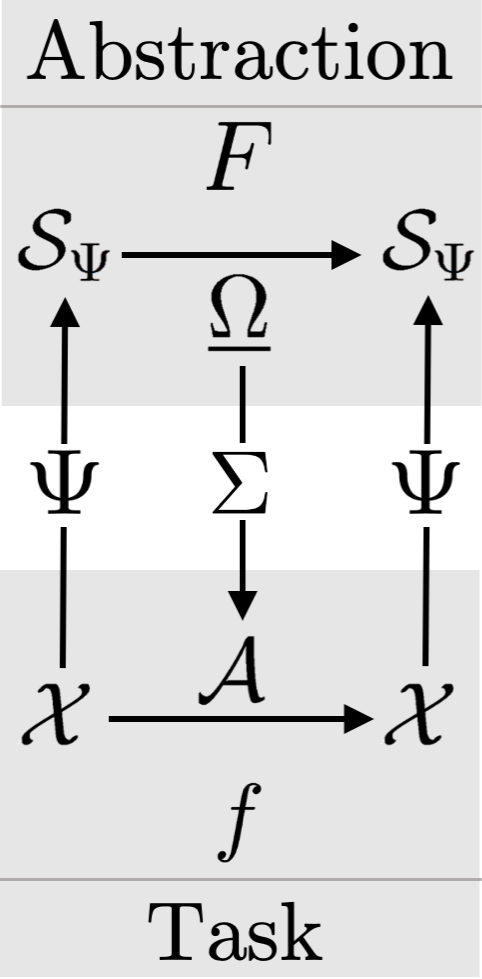}
\end{wrapfigure}

We adopt a very general definition of an abstraction~\cite{konidarisabstractions}: mappings from $\X$ and $\A$ to alternative state and action spaces. 
We first characterize an abstract state space $\S_\Psi$ and a transformation from states in $\X$ to abstract states.
Next, we describe an abstract action space $\ground{\operators}$ and an abstract transition model $F : \S_\Psi \times \ground{\operators} \to \S_\Psi$ that can be used to plan in the abstract space.
Finally, we define samplers $\samplers$ for refining abstract actions back into $\A$, i.e., actions that can be executed. See the diagram on the right for a summary.

\textbf{(1) An abstract state space.}
We use a set of predicates $\Psi$ (as defined in \secref{sec:problem}) to induce an abstract state space $\S_\Psi$.
Recalling that a ground atom $\ground{\psi}$ induces a classifier $c_{\ground{\psi}}$ over states $x \in \X$, we have:
\begin{defn}[Abstract state]
An \emph{abstract state} $s$ is the set of ground atoms under $\Psi$ that hold true in $x$: $$s = \textsc{Abstract}(x, \Psi) \triangleq \{ \ground{\psi} : c_{\ground{\psi}}(x) = \text{\emph{true}}, \forall \psi \in \Psi \}.$$
\end{defn}
The (discrete) abstract state space induced by $\Psi$ is denoted $\S_{\Psi}$.
Throughout this work, we use predicate sets $\Psi$ that are supersets of the given goal predicates $\Psi_G$.
However, only the goal predicates are given, and they alone are typically very limited; in \secref{sec:learning}, we will discuss how the agent can use data to \emph{invent predicates} that will make up the rest of $\Psi$. See \figref{fig:teaser} (first panel) for an example.

\textbf{(2) An abstract action space and abstract transition model.} We address both by having the agent learn \emph{operators}:

\begin{defn}[Operator]
An \emph{operator} is a tuple $\operator = \langle \params, \preconditions, \addeffects, \deleteeffects, \controllerspec \rangle$ where:
\begin{tightlist}
\item $\params$ is an ordered list of \emph{parameters}: variables with types drawn from the type set $\types$.
\item $\preconditions, \addeffects, \deleteeffects$ are \emph{preconditions}, \emph{add effects}, and \emph{delete effects}, each a set of lifted atoms over $\Psi$ and $\params$. 
\item $\controllerspec$ is a tuple $\langle C, \params_{\controllerspec} \rangle$ where $C((\type_1, \dots, \type_v), \Theta)$ is a controller and $\params_{\controllerspec}$ is an ordered list of \emph{controller arguments}, each a variable from $\params$. Furthermore, $|\params_{\controllerspec}| = v$, and each argument $i$ must be of the respective type $\type_i$.
\end{tightlist}
\end{defn}

We denote the set of operators as $\Omega$.
See \figref{fig:teaser} (second panel) for an example. Unlike in STRIPS~\cite{fikes1971strips}, our operators are augmented with controllers and controller arguments, which will allow us to connect to the task actions in \textbf{(3)} below. Now, given a task with object set $\O$, the set of all \emph{ground operators} defines our (discrete) abstract action space for a task:

\begin{defn}[Ground operator / abstract action]
A \emph{ground operator} $\ground{\operator} = \langle \operator, \delta \rangle$ is an operator $\operator$ and a substitution $\substitution : \params \to \O$ mapping parameters to objects.
We use $\ground{\preconditions}, \ground{\addeffects}, \ground{\deleteeffects}$, and $\ground{\params_{\controllerspec}}$ to denote the ground preconditions, ground add effects, ground delete effects, and ground controller arguments of $\ground{\operator}$, where variables in $\params$ are substituted with objects under $\substitution$.
\end{defn}

We denote the set of ground operators (the abstract action space) as $\ground{\operators}$. Together with the abstract state space $\S_\Psi$, the preconditions and effects of the operators induce an abstract transition model for a task:

\begin{defn}[Abstract transition model]
\label{def:absmodel}
The \emph{abstract transition model} induced by predicates $\Psi$ and operators $\operators$ is a partial function $F : \S_\Psi \times \ground{\operators} \to \S_\Psi$. $F(s, \ground{\operator})$ is only defined if $\ground{\operator}$ is \emph{applicable} in $s$:  $\ground{\preconditions} \subseteq s$. If defined, $F(s, \ground{\operator}) \triangleq (s - \ground{\deleteeffects}) \cup \ground{\addeffects}$.
\end{defn}

\textbf{(3) A mechanism for refining abstract actions into task actions.} A ground operator $\ground{\operator}$ induces a partially specified controller, $C((o_1, \ldots o_v), \Theta)$ with $(o_1, \ldots o_v) = \ground{\params_{\controllerspec}}$, where object arguments have been selected but continuous parameters $\Theta$ have not. To \emph{refine} this abstract action $\ground{\operator}$ into a task-level action $a = C((o_1, \ldots o_v), \theta)$, we use \emph{samplers}:

\begin{defn}[Sampler]
Each operator $\operator \in \operators$ is associated with a \emph{sampler} $\sampler : \X \times \O^{|\params|} \to \Delta(\Theta)$, where $\Delta(\Theta)$ is the space of distributions over $\Theta$, the continuous parameters of the operator's controller.
\end{defn}

\begin{defn}[Ground sampler]
For each ground operator $\ground{\operator} \in \ground{\operators}$, if $\ground{\operator} = \langle \operator, \delta \rangle$ and $\sampler$ is the sampler associated with $\operator$, then the $\emph{ground sampler}$ associated with $\ground{\operator}$ is a state-conditioned distribution $\ground{\sampler} : \X \to \Delta(\Theta)$, where $\ground{\sampler}(x) \triangleq \sampler(x, \delta(\params))$.
\end{defn}

We denote the set of samplers as $\samplers$. See \figref{fig:teaser} (third panel) for an example.

What connects the transition model $f$, abstract transition model $F$, and samplers $\samplers$? While previous works enforce the downward refinability property~\cite{marthi2007angelic,pasula2007learning,jetchev2013learning,konidaris2018skills}, it is important in robotics to be robust to violations of this property, since learned abstractions will typically lose critical geometric information.
Therefore, we only require our learned abstractions to satisfy the following \emph{weak semantics}:
for every ground operator $\ground{\operator}$ with partially specified controller $C((o_1, \dots, o_v), \Theta)$ and associated ground sampler $\ground{\sampler}$, there exists some $x \in \X$ and some $\theta$ in the support of $\ground{\sampler}(x)$ such that $F(s, \ground{\operator})$ is defined and equals $s'$, where $s = \textsc{Abstract}(x, \Psi)$, $a = C((o_1, \dots, o_v), \theta)$, and $s' = \textsc{Abstract}(f(x, a), \Psi)$.
Note that downward refinability~\cite{marthi2007angelic} makes a much stronger assumption: that this statement holds for \emph{every} $x \in \X$ where $F(s, \ground{\operator})$ is defined.

%% file: planning.tex
To use the components of an abstraction --- predicates $\Psi$, operators $\operators$, and samplers $\samplers$ --- for efficient planning, we build on \emph{bilevel} planning techniques~\cite{srivastava2014combined,garrett2021integrated}. We conduct an outer search over \emph{abstract plans} using the predicates and operators, and an inner search over refinements of an abstract plan into a task solution $\pi$ using the predicates and samplers.

\begin{defn}[Abstract plan]
An \emph{abstract plan} $\hat{\pi}$ for a task $\langle \O, x_0, g \rangle$ is a sequence of ground operators $(\ground{\operator}_1, \ldots, \ground{\operator}_n)$ such that applying the abstract transition model $s_i = F(s_{i-1}, \ground{\operator}_i)$ successively starting from $s_0 = \textsc{Abstract}(x_0, \Psi)$ results in a sequence of abstract states $(s_0, \ldots, s_n)$ that achieves the goal, i.e., $g \subseteq s_n$. This $(s_0, \ldots, s_n)$ is called the \emph{expected abstract state sequence}.
\end{defn}

Because downward refinability does not hold in our setting, an abstract plan $\hat{\pi}$ is \emph{not} guaranteed to be refinable into a solution $\pi$ for the task, which necessitates bilevel planning. We now describe the planning algorithm in detail.

The overall structure of the planner is outlined in \algref{alg:plan}. For the outer search that finds abstract plans $\hat{\pi}$, denoted \textsc{GenAbstractPlan} (\algrefshort{alg:plan}, Line 2), we leverage the STRIPS-style operators and predicates~\cite{fikes1971strips} to automatically derive a domain-independent heuristic popularized by the AI planning community, such as LMCut~\cite{helmert2009landmarks}. We use this heuristic to run an A$^*$ search over the abstract state space $\S_\Psi$ and abstract action space $\ground{\operators}$.
This A$^*$ search is used as a generator (hence the name \textsc{GenAbstractPlan}) of abstract plans $\hat{\pi}$, outputting one at a time\footnote{This usage of A$^*$ search as a generator is related to top-$k$ planning~\cite{katz-etal-icaps18,ren2021extended}. We experimented with off-the-shelf top-$k$ planners, but chose A$^*$ because it was faster in our domains. Note that the abstract plan generator is used heavily in learning (\secref{sec:learning}).}.
Parameter $n_{\text{abstract}}$ governs the maximum number of abstract plans that can be generated before the planner terminates with failure.

\input{planning_pseudocode}

For each abstract plan $\hat{\pi}$, we conduct an inner search that attempts to \textsc{Refine} (\algrefshort{alg:plan}, Line 3) it into a solution $\pi$ (a plan that achieves the goal under the transition model $f$). While various implementations of \textsc{Refine} are possible~\cite{chitnis2016guided}, we follow \citet{srivastava2014combined} and perform a backtracking search over the abstract actions $\ground{\operator}_i \in \hat{\pi}$. 
Recall that each $\ground{\operator}_i$ induces a partially specified controller $C_i((o_1, \dots, o_v)_i, \Theta_i)$ and has an associated ground sampler $\ground{\sampler}_i$.
To begin the search, we initialize an indexing variable $i$ to 1.
On each step of search, we sample continuous parameters $\theta_i \sim \ground{\sampler}_i(x_{i-1})$, which fully specify an action $a_i = C_i((o_1, \dots, o_v)_i, \theta_i)$. We then check whether $x_i = f(x_{i-1}, a_i)$ obeys the expected abstract state sequence, i.e., whether $s_i = \textsc{Abstract}(x_i, \Psi)$. If so, we continue on to $i \gets i + 1$. Otherwise, we repeat this step, sampling a new $\theta_i \sim \ground{\sampler}_i(x_{i-1})$. Parameter $n_{\text{samples}}$ governs the maximum number of times we invoke the sampler for a single value of $i$ before backtracking to $i \gets i - 1$. \textsc{Refine} succeeds if the goal $g$ holds when $i = |\hat{\pi}|$, and fails when $i$ backtracks to 0.

If \textsc{Refine} succeeds given a candidate $\hat{\pi}$, the planner terminates with success (\algrefshort{alg:plan}, Line 4) and returns the plan $\pi = (a_1, \ldots, a_{|\hat{\pi}|})$. Crucially, if \textsc{Refine} fails, we continue with \textsc{GenAbstractPlan} to generate the next candidate $\hat{\pi}$.
In the taxonomy of task and motion planners (TAMP), this approach is in the ``search-then-sample'' category~\cite{srivastava2014combined,dantam2016incremental,garrett2021integrated}.
As we have described it, this planner is \emph{not} probabilistically complete, because abstract plans are not revisited.
Extensions to ensure completeness are straightforward~\cite{chitnis2016guided}, but are not our focus in this work.

%% file: planning_pseudocode.tex
\begin{algorithm}[t]
  \SetAlgoNoEnd
  \DontPrintSemicolon
  \SetKwFunction{algo}{algo}\SetKwFunction{proc}{proc}
  \SetKwProg{myalg}{}{}{}
  \SetKwProg{myproc}{Subroutine}{}{}
  \SetKw{Continue}{continue}
  \SetKw{Break}{break}
  \SetKw{Return}{return}
  \SetKwFor{For}{for}{}{}
\myalg{\textsc{Plan(}$x_0$, $g$, $\Psi$, $\Omega$, $\Sigma$ \textsc{)}}{
    \tcp{\footnotesize Parameters: $n_{\text{abstract}}$, $n_{\text{samples}}$.}
    \nl $s_0 \gets \textsc{Abstract}(x_0, \Psi)$\;
    \nl \For{$\hat{\pi}$ in \textsc{GenAbstractPlan}($s_0$, $g$, $\Omega$, $n_{\text{abstract}}$)}
    {
     
    \nl \If{$\pi \sim $ \textsc{Refine}($\hat{\pi}$, $x_0$, $\Psi$, $\Sigma$, $n_{\text{samples}}$)} {
        \nl \Return $\pi$\;
    }
    }
    }\;
\caption{\small{Pseudocode for our bilevel planning algorithm. The inputs are an initial state $x_0$, goal $g$, predicates $\Psi$, operators $\operators$, and samplers $\samplers$; the output is a plan $\pi$. An outer loop runs \textsc{GenAbstractPlan}, which generates plans in the abstract state and action spaces. An inner loop runs \textsc{Refine}, which attempts to refine each abstract plan $\hat{\pi}$ into a plan $\pi$. If \textsc{Refine} succeeds, then the found plan $\pi$ is returned as the solution; if \textsc{Refine} fails, then \textsc{GenAbstractPlan} continues.}}
\label{alg:plan}
\end{algorithm}

%% file: learning.tex
To use bilevel planning at evaluation time, we must learn predicates, operators, and samplers at training time.
We use the methods of~\citet{chitnis2021learning} for operator learning and sampler learning; see \secref{subsec:oplearning} and \secref{subsec:samplerlearning} for descriptions.
For what follows, it is important to understand that operator learning is fast ($O(|\D|)$), but sampler learning is slow, and both require a given set of predicates.
Our main contribution is a method for predicate invention that precedes operator and sampler learning in the training pipeline.

\input{surrogate_pseudocode}

Inspired by prior work~\cite{bonet2019learning,loula2019discovering,curtis2021discovering}, we approach the predicate invention problem from a program synthesis perspective~\cite{stahl1993predicate,lavrac1994inductive,cropper2016learning,ellis2020dreamcoder}.
First, we define a compact representation of an infinite space of predicates in the form of a \emph{grammar}.
We then enumerate a large pool of \emph{candidate predicates} from this grammar, with simpler candidates enumerated first.
Next, we perform a \emph{local search} over subsets of candidates, with the aim of identifying a good final subset to use as $\Psi$.
The crucial question in this step is: what \emph{objective function} should we use to guide the search over candidate predicate sets?

\begin{figure*}[t]
  \centering
    \noindent
    \includegraphics[width=\textwidth]{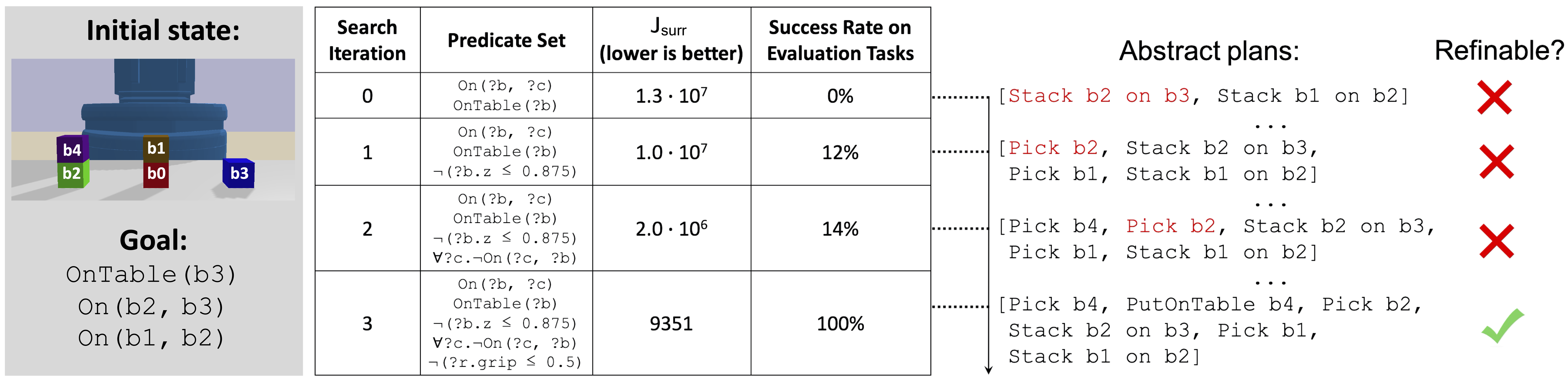}
    \caption{\small{\textbf{Predicate invention via hill climbing}. (Left) An example task in Blocks. (Middle) Hill climbing over predicate sets, starting with the goal predicates $\Psi_G$. On each iteration, the single predicate that improves $J_{\text{surr}}$ the most is added to the set. The rightmost table column shows success rates on held-out evaluation tasks.
    Each iteration of hill climbing adds a predicate that causes all abstract plans above the dotted line to be pruned from consideration. At iteration 0, the robot believes it can achieve the goal by simply stacking \texttt{b2} on \texttt{b3} and \texttt{b1} on \texttt{b2}, even though it hasn't picked up either block. The first step of this abstract plan (shown in red) is thus unrefinable. At iteration 1, a predicate with the intuitive meaning \texttt{Holding} is added, which makes the A$^*$ only consider abstract plans that pick up blocks before stacking them. Still, the abstract plan shown is unrefinable on the first step because \texttt{b4} is obstructing \texttt{b2} in the initial state. At iteration 2, a predicate with the intuitive meaning \texttt{NothingAbove} is added, which allows the agent to realize that it must move \texttt{b4} out of the way if it wants to pick up \texttt{b2}. This plan is still unrefinable, though: the second step fails, because the abstraction still does not recognize that the robot cannot be holding two blocks simultaneously. Finally, at iteration 3, a predicate with the intuitive meaning \texttt{HandEmpty} is added, and planning succeeds.
    }}
  \label{fig:hill_climbing}
\end{figure*}

\subsection{Scoring a Candidate Predicate Set}
\label{subsubsec:scorefunc}

Ultimately, we want to find a set of predicates $\Psi$ that will lead to efficient planning, after we use the predicates to learn operators $\operators$ and samplers $\samplers$.
I.e., our real objective is: $$J_{\text{real}}(\Psi) \triangleq \mathbb{E}_{(\O, x_0, g) \sim \T}[\textsc{Time}(\textsc{Plan}(x_0, g, \Psi, \operators, \samplers))],$$
where $\operators$ and $\samplers$ are learned using $\Psi$ as we described in Sections \ref{subsec:oplearning} and \ref{subsec:samplerlearning}, \textsc{Plan} is the algorithm described in \secref{sec:planning}, and $\textsc{Time}(\cdot)$ measures the time that \textsc{Plan} takes to find a solution\footnote{If no plan can be found (e.g., a task is infeasible under the abstraction), \textsc{Time} would return a large constant representing a timeout.}.
However, we need an objective that can be used to guide a \emph{search} over candidate predicate sets, meaning the objective must be evaluated many times. 
$J_{\text{real}}$ is far too expensive for this, due to two speed bottlenecks: sampler learning, which involves training several neural networks; and the repeated calls to \textsc{Refine} from within \textsc{Plan}, which each perform backtracking search to refine an abstract plan.
To overcome this intractability, we will use a \emph{surrogate objective} $J_{\text{surr}}$ that is cheaper to evaluate than $J_{\text{real}}$, but that approximately preserves the ordering over predicate sets, i.e., $J_{\text{surr}}(\Psi) < J_{\text{surr}}(\Psi') \iff J_{\text{real}}(\Psi) < J_{\text{real}}(\Psi')$.



We propose a surrogate objective that uses the demonstrations $\D$ to \emph{estimate} the time it would take to solve the training tasks under the abstraction induced by a candidate predicate set $\Psi$, without using samplers or doing refinement.
Recalling that $\D$ has one demonstration $\pi^*$ for each training task $\langle \O, x_0, g \rangle$, the objective is defined as follows:
$$J_{\text{surr}}(\Psi) \triangleq \frac{1}{|\D|} \sum_{(\O, x_0, g, \pi^*) \in \D}[\textsc{ETPT}(x_0, g, \Psi, \operators, \pi^*)],$$
where \textsc{ETPT} abbreviates Estimate Total Planning Time (\algref{alg:surrogate}).
\textsc{ETPT} uses the candidate predicates and induced operators to perform the first part of bilevel planning: A$^*$ search over abstract plans.
However, for each generated abstract plan, rather than learning samplers and calling \textsc{Refine}, we use the available demonstrations to estimate the probability that refinement \emph{would} succeed if we \emph{were} to learn samplers and call \textsc{Refine}.
Since bilevel planning terminates upon the successful refinement of an abstract plan, we can use these probabilities to approximate the total expected planning time.
We now describe these steps in detail.

\subsubsection{Estimating Refinement Probability}
As \textsc{ETPT} generates abstract plans, it estimates the probability that each is the \emph{first} to be \emph{successfully refined}.
An abstract plan is first if no previous plan was already refined.
This happens with probability $p_\text{continue}$, which is initialized to $1$ (Line 2) and updated after each iteration (Line 9).
We then use a helper function \textsc{EstimateRefineProb} to approximate the probability of successfully refining the given abstract plan, if we were to learn samplers $\samplers$ and then call \textsc{Refine}.
We use the following implementation: $$\textsc{EstimateRefineProb}(\hat{\pi}, \pi^*) \triangleq (1 - \epsilon)\epsilon^{|\textsc{Cost}(\hat{\pi}) - \textsc{Cost}(\pi^*)|}.$$
Here, $\epsilon > 0$ is a small constant ($10^{-5}$ in our experiments), and $\textsc{Cost}(\cdot)$ is in our case simply the number of actions in the plan, due to unitary costs.
The intuition for this geometric distribution is as follows. Since the demonstration $\pi^*$ is assumed to be near-optimal, an abstract plan $\hat{\pi}$ that is cheaper than $\pi^*$ should look suspicious; if such a $\hat{\pi}$ were refinable, then the demonstrator would have likely used it to produce a better demonstration. If $\hat{\pi}$ is more expensive than $\pi^*$, then even though this abstraction would eventually produce a refinable abstract plan, it may take a long time for the outer loop of the planner, \textsc{GenAbstractPlan}, to get to it (\secref{sec:planning}).
We note that this scheme for estimating refinability is surprisingly minimal, in that it needs only the cost of each demonstration rather than its contents.

\subsubsection{Estimating Time}
To approximate the total planning time, ETPT estimates the time required for each generated abstract plan, conditioned on its successful refinement, and then uses the refinement probabilities to compute the total expectation.
The time estimate is maintained in $t_\text{expected}$, initialized to 0 (Line 3).
To update $t_\text{expected}$ on each abstract plan (Lines 7-8), we use a helper function \textsc{EstimateTime}, which sums together estimates of the \emph{abstract search time} and of the \emph{refinement time}.
Since we are running abstract search, we could exactly measure its time; however, to avoid noise due to CPU speed, we instead use the cumulative number of nodes created by the A$^*$ search.
To estimate refinement time, recall that \textsc{Refine} performs a backtracking search, and so over many calls to \textsc{Refine}, the potentially several that fail will dominate the one or zero that succeed. Therefore, we estimate refinement time as a large constant ($10^3$ in our experiments) that captures the average cost of an exhaustive backtracking search.
Finally, we use a large constant $t_{\text{upper}}$ ($10^5$ in our experiments) to penalize in the case where no abstract plan succeeds (Line 10).

What is the ideal choice for $n_{\text{abstract}}$, the maximum number of abstract plans to consider within \textsc{ETPT}?
From an efficiency perspective, $n_{\text{abstract}}=1$ is ideal, but otherwise, it is not obvious whether to prefer the value of $n_{\text{abstract}}$ that will eventually be used with \textsc{Plan} at evaluation time, or to instead prefer $n_{\text{abstract}}=\infty$.
On one hand, we want \textsc{ETPT} to be as much of a mirror image of \textsc{Plan} as possible; on the other hand, some experimentation we conducted suggests that a larger value of $n_{\text{abstract}}$ can smooth the objective landscape, which makes search easier.
In practice, it may be advisable to treat $n_{\text{abstract}}$ as a hyperparameter.

In summary, our surrogate objective $J_{\text{surr}}$ calculates and combines two characteristics of a candidate predicate set $\Psi$: (1) abstract plan cost ``error,'' i.e., $|\textsc{Cost}(\hat{\pi}) - \textsc{Cost}(\pi^*)|$; and (2) abstract planning time, i.e., number of nodes created during A$^*$.
The first feature uses only the costs of the demonstrated plans, while the second feature does not use the demonstrated plans at all.
In \appref{app:results}, we conduct an empirical analysis to further unpack the contribution of these two features to the overall surrogate objective, finding them to be helpful together but insufficient  individually.

\subsection{Local Search over Candidate Predicate Sets}
\label{subsubsec:localsearch}

With our surrogate objective $J_{\text{surr}}$ established, we turn to the question of how to best optimize it.
We use a simple hill-climbing search, initialized with $\Psi_0 \gets \Psi_G$, and adding a single new predicate $\psi$ from the pool on each step $i$: $$\Psi_{i+1} \gets \argmin_{\psi \not\in \Psi_i} J_{\text{surr}}(\Psi_i \cup \{\psi\}).$$
We repeat until no improvement can be found, and use the last predicate set as our final $\Psi$.
See \figref{fig:hill_climbing} for an example taken from our experiments in the Blocks environment.

\subsubsection{Designing a Grammar of Predicates}
\label{subsubsec:grammar}

Designing a grammar of predicates can be difficult, since there is a tradeoff between the expressivity of the grammar and the practicality of searching over it.
For our experiments, we found that a simple grammar similar to that of \citet{pasula2007learning} suffices, which includes single-feature inequalities, logical negation, and universal quantification.
See \appref{app:grammar} for a full description and \figref{fig:teaser} and \appref{app:results} for examples.

The costs accumulated over the production rules lead us to a final cost associated with each predicate $\psi$, denoted $\textsc{pen}(\psi)$, where a higher cost represents a predicate with higher complexity.
We use the costs to regularize $J_{\text{surr}}$ during local search, with a weight small enough to primarily prevent the addition of ``neutral'' predicates that neither harm nor hurt $J_{\text{surr}}$.
The regularization term is $J_{\text{reg}}(\Psi) \triangleq w_{\text{reg}} \sum_{\psi \in \Psi} \textsc{pen}(\psi)$, where $w_{\text{reg}} = 10^{-4}$ in our experiments.
To generate our candidate predicate set for local search, we enumerate $n_{\text{grammar}}$ (200 in experiments) predicates from the grammar, in order of increasing cost.

%% file: surrogate_pseudocode.tex
\begin{algorithm}[t]
  \SetAlgoNoEnd
  \DontPrintSemicolon
  \SetKwFunction{algo}{algo}\SetKwFunction{proc}{proc}
  \SetKwProg{myalg}{}{}{}
  \SetKwProg{myproc}{Subroutine}{}{}
  \SetKw{Continue}{continue}
  \SetKw{Break}{break}
  \SetKw{Return}{return}
  \SetKwFor{For}{for}{}{}
  \myalg{\textsc{ETPT(}{\color{blue} $x_0$, $g$, $\Psi$, $\Omega$}, $\pi^*$\textsc{)}}{
  \tcp{\footnotesize Note: does not take in samplers!}
  \tcp{\footnotesize Parameters: {\color{blue} $n_{\text{abstract}}$},  $t_{\text{upper}}$.}
    \nl {\color{blue} $s_0 \gets \textsc{Abstract}(x_0, \Psi)$\;}
    \nl  $p_{\text{continue}} \gets 1.0$\;
    \nl  $t_{\text{expected}} \gets 0.0$\;
    \nl \For{{\color{blue} $\hat{\pi}$ in \textsc{GenAbstractPlan}($s_0$, $g$, $\Omega$, $n_{\text{abstract}}$)}}
    {
    \nl $p_{\text{refined}} \gets$ \textsc{EstimateRefineProb($\hat{\pi}$, $\pi^*$)}\;
    \nl  $p_{\text{terminate}} \gets p_{\text{continue}} \cdot p_{\text{refined}}$\;
    \nl  $t_{\text{iter}} \gets$ \textsc{EstimateTime({\color{blue}$\hat{\pi}$, $x_0$, $\Psi$}, $\operators$}) \;
    \nl  $t_{\text{expected}} \gets t_{\text{expected}} + p_{\text{terminate}}\cdot t_{\text{iter}}$ \;
    \nl $p_{\text{continue}} \gets p_{\text{continue}}\cdot(1 - p_{\text{refined}})$}
    \nl $t_{\text{expected}} \gets t_{\text{expected}}  + p_{\text{continue}}\cdot t_{\text{upper}}  $\;
    \nl \textbf{return} $t_{\text{expected}}$}\;
\caption{\small{Pseudocode for Estimate Total Planning Time in our predicate invention surrogate objective. Commonalities with \algref{alg:plan} are shown in blue. See \secref{sec:learning} for details.}}
\vspace{-1em}
\label{alg:surrogate}
\end{algorithm}

%% file: experiments.tex
\begin{figure*}[t]
  \centering
    \noindent
    \includegraphics[width=\textwidth]{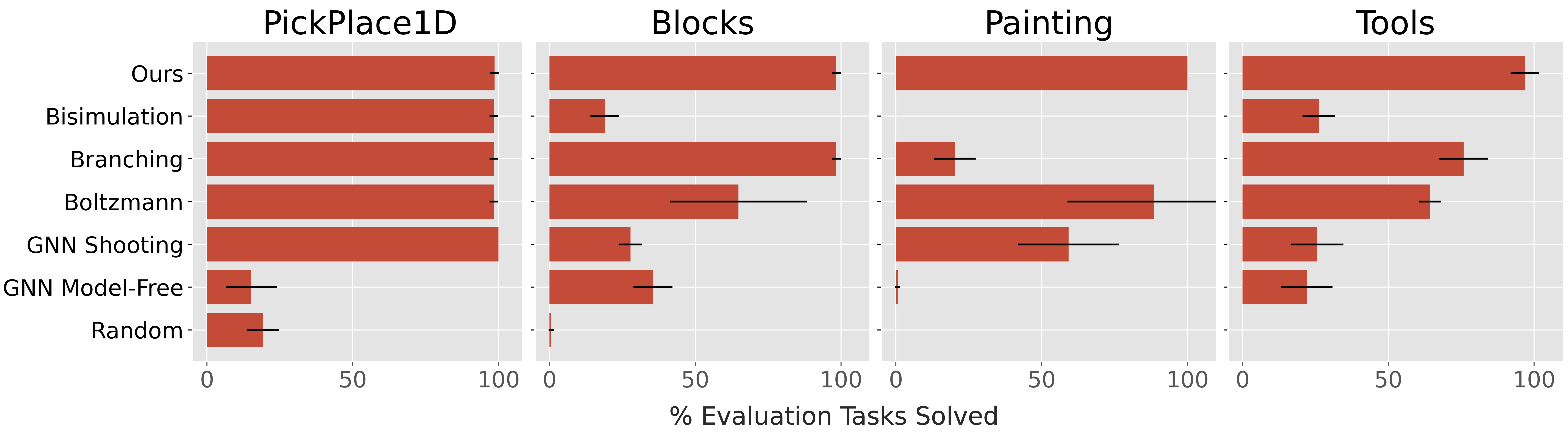}
    \caption{\small{\textbf{Ours versus baselines.} Percentage of 50 evaluation tasks solved under a 10-second timeout, for all four environments. All results are averaged over 10 seeds. Black bars denote standard deviations.
    Learning times and additional metrics are reported in \appref{app:results}.}}
  \label{fig:barplots}
\end{figure*}

\input{ablations_table}


Our experiments are designed to answer the following questions:
\textbf{(Q1)} To what extent do our learned abstractions help both the effectiveness and the efficiency of planning, and how do they compare to abstractions learned using other objective functions?
\textbf{(Q2)} How do our learned state abstractions compare in performance to manually designed state abstractions?
\textbf{(Q3)} How data-efficient is learning, with respect to the number of demonstrations?
\textbf{(Q4)} Do our abstractions vary as we change the planner configuration, and if so, how?

\subsubsection{Experimental Setup}
We evaluate 10 methods across four robotic planning environments.
All results are averaged over 10 random seeds.
For each seed, we sample a set of 50 \emph{evaluation tasks} that involve more objects and harder goals than were seen at training.
Demonstrations are collected by bilevel planning with manually defined abstractions (see Manual method below).
Planning is always limited to a 10-second timeout.
See \appref{app:additional-experimental-details} for additional details.

\subsubsection{Environments} We now briefly describe the environments, with further details in \appref{app:details}. The first three environments were established in prior work by ~\citet{silver2021learning}, but in that work, all predicates were manually defined; we use the same predicates in the Manual baseline.
\begin{tightlist}
\item \textbf{PickPlace1D.} A robot must pick blocks and place them onto target regions along a table surface. All pick and place poses are in a 1D line. 
Evaluation tasks require 1-4 actions to solve.
\item \textbf{Blocks.} A robot in 3D must interact with blocks on a table to assemble them into towers. This is a robotic version of the classic blocks world domain.
Evaluation tasks require 2-20 actions to solve.
\item \textbf{Painting.} A robot in 3D must pick, wash, dry, paint, and place widgets into either a box or a shelf.
Evaluation tasks require 11-25 actions to solve.
\item \textbf{Tools.} A robot operating on a 2D table surface must assemble contraptions with screws, nails, and bolts, using a provided set of screwdrivers, hammers, and wrenches respectively. This environment has physical constraints that cannot be modeled by our predicate grammar.
Evaluation tasks require 7-20 actions to solve.
\end{tightlist}

\subsubsection{Methods} We evaluate our method, six baselines, a manually designed state abstraction, and two ablations.
Note that the Bisimulation, Branching, Boltzmann, and Manual baselines differ from Ours only in predicate learning.
\begin{tightlist}
\item \textbf{Ours.} Our main approach.
\item \textbf{Bisimulation.} A baseline that learns abstractions by approximately optimizing the \emph{bisimulation criteria}~\cite{givan2003equivalence}, as in prior work~\cite{curtis2021discovering}. Specifically, this baseline learns abstractions that minimize the number of transitions in the demonstrations where the abstract transition model $F$ is applicable but makes a misprediction about the next abstract state. Note that because goal predicates are given, goal distinguishability is satisfied under any abstraction.
\item \textbf{Branching.} A baseline that learns abstractions by optimizing the \emph{branching factor} of planning. Specifically, this baseline learns predicates that minimize the number of applicable operators over demonstration states.
\item \textbf{Boltzmann.} A baseline that assumes the demonstrator is acting \emph{noisily rationally} under (unknown) optimal abstractions~\cite{baker2009action}. For any candidate abstraction, we compute the likelihood of the demonstration under a Boltzmann policy using the planning heuristic as a surrogate for the true cost-to-go.
\item \textbf{GNN Shooting.} A baseline that trains a graph neural network~\cite{battaglia2018relational} policy. This GNN takes in the current state $x$, abstract state $s$, and goal $g$. It outputs an action $a$, via a one-hot vector over $\C$ corresponding to which controller to execute, one-hot vectors over all objects at each discrete argument position, and a vector of continuous arguments. We train the GNN using behavior cloning on the data $\D$. At evaluation time, we sample trajectories by treating the outputted continuous arguments as the mean of a Gaussian with fixed variance. We use the transition model $f$ to check if the goal is achieved, and repeat until the planning timeout is reached.
\item \textbf{GNN Model-Free.} A baseline that uses the same GNN, but directly executes the policy instead of shooting.
\item \textbf{Random.} A baseline that simply executes a random controller with random arguments on each step. No learning.
\item \textbf{Manual.} An oracle approach that plans with manually designed predicates for each environment.
\item \textbf{Down Eval.} An ablation of Ours that uses $n_{\text{abstract}} = 1$ during evaluation only, in \textsc{Plan} (\algref{alg:plan}).
\item \textbf{No Invent.} An ablation of Ours that uses $\Psi = \Psi_G$, i.e., only goal predicates are used for the state abstraction.
\end{tightlist}

\subsubsection{Results and Discussion}

We provide real examples of learned predicates and operators for all environments in \appref{app:results}.
\figref{fig:barplots} shows that our method solves many more held-out tasks within the timeout than the baselines.
A major reason for this performance gap is that our surrogate objective $J_{\text{surr}}$ explicitly approximates the efficiency of planning.
The lackluster performance of the bisimulation baseline is especially notable because of its prevalence in the literature~\cite{pasula2007learning, jetchev2013learning,bonet2019learning,curtis2021discovering}.
We examined its failure modes more closely and found that it consistently selects good predicates, but not \emph{enough} of them.
This is because requiring the operators to be a perfect predictive model in the abstract spaces is often not enough to ensure good planning performance. For example, in the Blocks environment, the goal predicates together with the predicate \texttt{Holding(?block)} are enough to satisfy bisimulation on our data, while other predicates like \texttt{Clear(?block)} and \texttt{HandEmpty()} are useful from a planning perspective.
Examining the GNN baselines, we see that while shooting is beneficial versus using the GNN model-free, the performance is generally far worse than Ours.
Additional experimentation we conducted suggests that the GNN gets better with around an order of magnitude more data.

\begin{wrapfigure}[10]{r}{0.29\textwidth}
\vspace{-1.5em}
\hspace{-1em}
\includegraphics[width=0.29\textwidth]{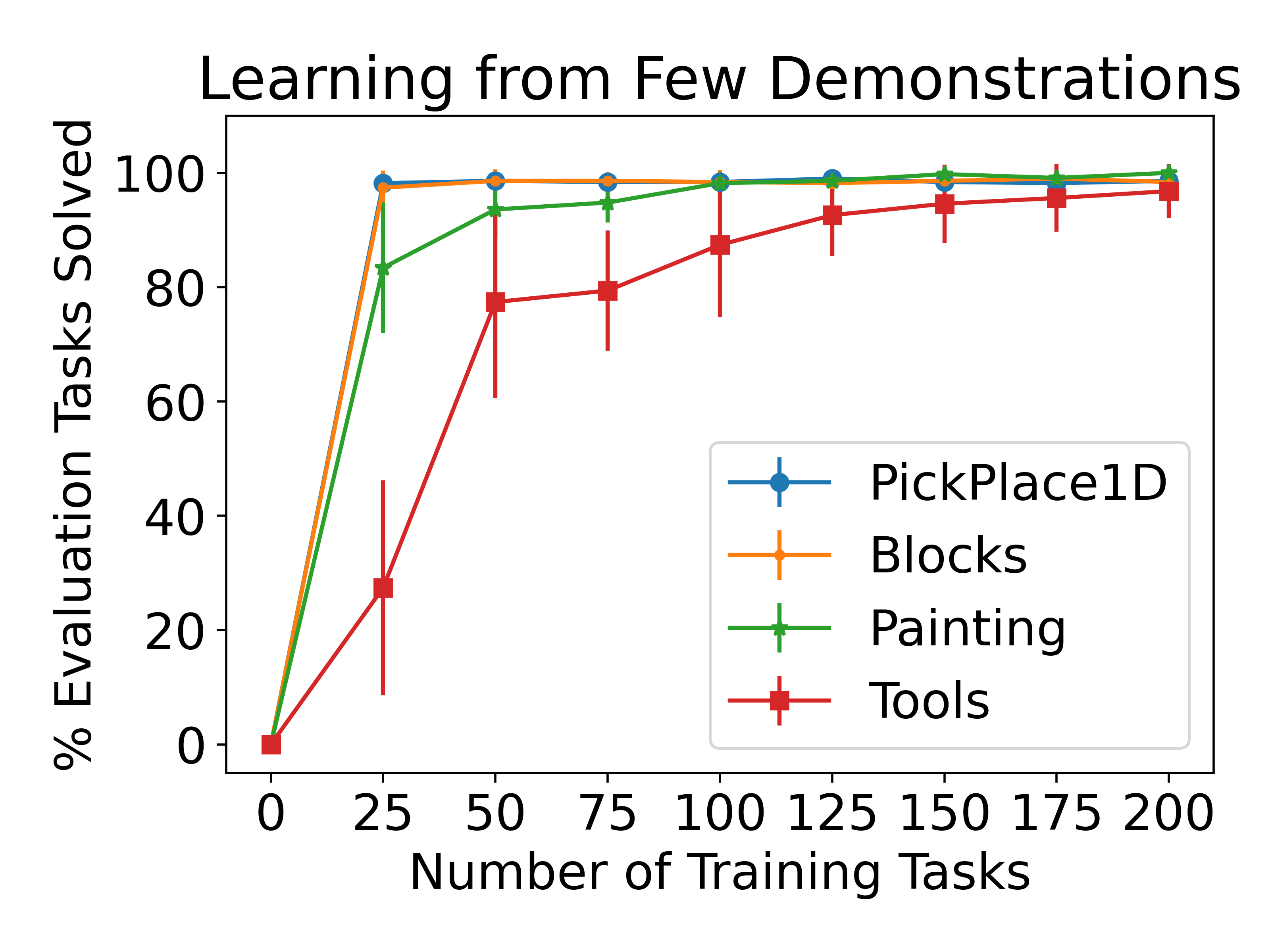}
\end{wrapfigure}

The figure on the right illustrates the data efficiency of Ours. Each point shows a mean over 10 seeds, with standard deviations shown as vertical bars.
We often obtain very good evaluation performance within just 50 demonstrations.

In \tabref{tab:ablations}, the results for No Invent show that, as expected, the goal predicates alone are completely insufficient for most tasks.
Comparing Ours to Down Eval shows that assuming downward refinability at evaluation time works for PickPlace1D, Blocks, and Painting, but not for Tools.
We also find that the learned predicates (Ours) are on par with, and sometimes \emph{better than}, hand-designed predicates (Manual).
For instance, consider PickPlace1D, where the learned predicates are 7.5x better.
The manually designed predicates were \texttt{Held(?block)} and \texttt{HandEmpty()}, and the always-given goal predicate \texttt{Covers(?block, ?target)}.
In addition to inventing two predicates that are equivalent to \texttt{Held} and \texttt{HandEmpty}, Ours invented two more: $\texttt{P3(?block)} \triangleq \forall \texttt{?t}\ .\ \lnot \texttt{Covers(?block, ?t)}$, and $\texttt{P4(?target)} \triangleq \forall \texttt{?b}\ .\ \lnot \texttt{Covers(?b, ?target)}$. 
Intuitively, \texttt{P3} means ``the given block is not on any target,'' while \texttt{P4} means ``the given target is clear.'' 
\texttt{P3} gets used in an operator precondition for picking, which reduces the branching factor of abstract search.
This precondition is sensible because there is no use in moving a block once it is already on its target.
\texttt{P4} prevents considering non-refinable abstract plans that ``park'' objects on targets that must be covered by other objects.

In \appref{app:more-blocks-explanations}, we describe an additional experiment where we vary the AI planning heuristic used in abstract search.
We analyze a case in Blocks where variation in the invented predicates appears inconsequential upon initial inspection, but actually has substantial impact on planning efficiency.
This result underscores the benefit of using a surrogate objective for predicate invention that is sensitive to downstream planning efficiency.

%% file: ablations_table.tex
\begin{table*}
	\centering
	\resizebox{0.85\textwidth}{!}{%
    \begin{tabular}{| l | p{0.75cm} | p{0.75cm} | p{0.75cm} || p{0.75cm} | p{0.75cm} | p{0.75cm} || p{0.75cm} | p{0.75cm} | p{0.75cm} || p{0.75cm} | p{0.75cm} | p{0.75cm} | }
	\hline
	\multicolumn{1}{|c|}{} &\multicolumn{3}{c||}{\bf{Ours}} &
	\multicolumn{3}{c||}{\bf{Manual}} &
	\multicolumn{3}{c||}{\bf{Down Eval}} &
	\multicolumn{3}{c|}{\bf{No Invent}} \\
	\hline
	\bf{Environment} &
	{Succ} & {Node} & {Time}&
	{Succ} & {Node} & {Time}&
	{Succ} & {Node} & {Time}&
	{Succ} & {Node} & {Time}\\
	\hline
	PickPlace1D & 98.6 & 4.8 & 0.006 & 98.4 & 6.5 & 0.045 & 98.6 & 4.8 & 0.008 & 39.6 & 14.1 & 1.369 \\
	Blocks & 98.4 & 2949 & 0.296 & 98.6 & 2941 & 0.251 & 98.2 & 2949 & 0.318 & 3.2 & 427.7 & 1.235 \\
	Painting & 100.0 & 501.8 & 0.470 & 99.6 & 2608 & 0.464 & 98.8 & 489.0 & 0.208 & 0.0 & \;\;\;-- & \;\;\;-- \\
	Tools & 96.8 & 1897 & 0.457 & 100.0 & 4771 & 0.491 & 42.8 & 152.5 & 0.060 & 0.0 & \;\;\;-- & \;\;\;-- \\\hline
	\end{tabular}}
    
	\caption{\small{\textbf{Ours versus Manual and ablations.} Percentage of 50 evaluation tasks solved under a 10-second timeout (Succ), number of nodes created during \textsc{GenAbstractPlan} (Node), and wall-clock planning time in seconds (Time). All results are averaged over 10 seeds. The Node and Time columns average over \emph{solved tasks only}. Standard deviations are provided in \appref{app:results}.}}
	\label{tab:ablations}
\end{table*}

%% file: related_work.tex
Our work continues a long line of research on learning state abstractions for decision-making \cite{bertsekas1988adaptive,andre2002state,jong2005state,li2006towards,abel2017near,zhang2020learning}.
Most relevant are works that learn symbolic abstractions compatible with AI planners~\cite{lang2012exploration,jetchev2013learning,ugur2015bottom,asai2018classical,bonet2019learning,asai2020learning,ahmetoglu2020deepsym,umili2021learning}.
Our work is particularly influenced by \citet{pasula2007learning}, who use search through a concept language to invent symbolic state and action abstractions, and \citet{konidaris2018skills}, who discover symbolic abstractions by leveraging the initiation and termination sets of options that satisfy an abstract subgoal property.
The objectives used in these prior works are based on variations of auto-encoding, prediction error, or bisimulation, which stem from the perspective that the abstractions should \emph{replace} planning in the original transition space, rather than \emph{guide} it.

Recent works have also considered learning abstractions for multi-level planning, like those in the task and motion planning (TAMP)~\cite{gravot2005asymov,garrett2021integrated} and hierarchical planning~\cite{bercher2019survey} literature.
Some of these efforts consider learning symbolic action abstractions~\cite{zhuo2009learning,nguyen2017automated,silver2021learning,aineto2022comprehensive} or refinement strategies~\cite{chitnis2016guided,mandalika2019generalized,chitnis2019learning,wang2021learning,chitnis2021learning,ortiz2022structured}; our operator and sampler learning methods take inspiration from these prior works.
Recent efforts by \citet{loula2019discovering} and \citet{curtis2021discovering} consider learning both state and action abstractions for TAMP, like we do \cite{loula2019discovering,loula2020learning,curtis2021discovering}.
The main distinguishing feature of our work is that our abstraction learning framework explicitly optimizes an objective that considers downstream planning efficiency.

%% file: conclusion.tex
In this paper, we have described a method for learning predicates that are explicitly optimized for efficient bilevel planning.
Key areas for future work include (1) learning better abstractions from even fewer demonstrations by performing active learning to gather more data online; (2) expanding the expressivity of the grammar to learn more sophisticated predicates; (3) applying these ideas to partially observed planning problems; and (4) learning the controllers that we assumed given in this work.

For (1), we hope to investigate how relational exploration algorithms~\cite{chitnis2020glib} might be useful as a mechanism for an agent to decide what actions to execute, toward the goal of building better state and action abstractions.
For (2), we can take inspiration from program synthesis, especially methods that can learn programs with continuous parameters~\cite{ellis2020dreamcoder}.
For (3) we could draw insights from recent advances in task and motion planning in the partially observed setting~\cite{garrett2020online}.
Finally, for (4), we recently proposed a method for learning controllers from demonstrations assuming known predicates~\cite{silver2022learning}.
If we can remove the latter assumption, we will have a complete pipeline for learning predicates, operators, samplers, and controllers for bilevel planning.

%% file: acknowledgements.tex
We gratefully acknowledge support from NSF grant 2214177; from AFOSR grant FA9550-22-1-0249; from ONR MURI grant N00014-22-1-2740; from the MIT-IBM Watson Lab; and from the MIT Quest for Intelligence.
Tom, Nishanth and Willie are supported by NSF Graduate Research Fellowships.
We thank Michael Katz, Christian Muise, Aidan Curtis, Jiayuan Mao, Zhutian Yang, and Amber Li for helpful comments on an earlier draft.
We also thank Yichao Liang for fixing an error in Algorithm 2.

%% file: appendix.tex
\subsection{Operator Learning}
\label{subsec:oplearning}
Here we describe how to learn operators $\operators$, assuming that the full set of predicates $\Psi$ is already learned~\cite{chitnis2021learning}.
This method makes two restrictions on the representation that together lead to very efficient operator learning (linear time in the number of transitions in $\D$).
First, for each $\controllerspec$ and each possible effect set pair $(\addeffects, \deleteeffects)$, there is at most one operator with that $(\controllerspec, \addeffects, \deleteeffects)$.
This restriction makes it impossible to learn multiple operators with different preconditions for the same controller and effect sets.
Second, each parameter in $\params$ must appear in $\params_{\controllerspec}, \addeffects$, or $\deleteeffects$.
This restriction prevents modeling ``indirect effects,'' where some object impacts the execution of a controller without its own state being changed.
Though these two restrictions are limiting, we are willing to accept them because predicate invention can compensate.
For example, an invented predicate can quantify out an object that does not appear in the controller or the effects, to capture indirect effects.

With these restrictions established, we learn operators from our demonstrations $\D$ and predicates $\Psi$ in three steps. Note that each demonstration can be expressed as a sequence of transitions $\{(x, a, x')\}$, with $x, x' \in \X$ and $a \in \A$.
First, we use $\Psi$ to \textsc{Abstract} all states $x, x'$ in the demonstrations $\D$, creating a dataset of transitions $\{(s, a, s')\}$ with $s, s' \in \S_\Psi$.
Next, we partition these transitions using the following equivalence relation: $(s_1, a_1, s_1') \equiv (s_2, a_2, s_2')$ if the effects and partially specified controllers \emph{unify}, that is, if there exists a mapping between the objects such that $a_1$, $(s_1 - s_1')$, and $(s_1' - s_1)$ are equivalent to $a_2$, $(s_2 - s_2')$, and $(s_2' - s_2)$ respectively.
This partitioning step automatically determines the number of operators that will ultimately be learned: each equivalence class will induce one operator.
Furthermore, the parameters $\params$, controller tuple $\controllerspec$, and effects $(\addeffects, \deleteeffects)$ of the operators can now be established as follows.
For each equivalence class, we create $\params$ by selecting an arbitrary transition $(s, a, s')$ and replacing each object that appears in the controller or effects with a variable of the same type. This further induces a substitution $\delta: \params \to \O$ for the objects $\O$ in this transition; the $\controllerspec$, $\addeffects$, and $\deleteeffects$ are then created by applying $\delta$ to $a, (s' - s)$, and $(s - s')$ respectively.
By construction, for all other transitions $\tau$ in the same equivalence class, there exists an injective substitution $\delta_\tau$ under which the controller arguments and effects are equivalent to the newly created $\controllerspec, \addeffects$, and $\deleteeffects$.
We use these substitutions for the third and final step of operator learning: precondition learning.
For this, we perform an intersection over all abstract states in each equivalence class~\cite{bonet2019learning,chitnis2021learning,curtis2021discovering}: $\preconditions \gets \bigcap_{\tau=(s, \cdot, \cdot)} \delta^{-1}_\tau(s),$
where $\delta^{-1}_\tau(s)$  substitutes all occurrences of the objects in $s$ with the parameters in $\params$ following an inversion of $\delta_\tau$, and discards any atoms involving objects that are not in the image of $\delta_\tau$.
By this construction, only the parameters in $\params$ will be involved in $\preconditions$, as desired.
With $\params, \preconditions, \addeffects, \deleteeffects$, and $\controllerspec$ now established for each equivalence class, we have completed the operators $\operators$.

\textbf{Soundness.} We note that for any predicates $\Psi$, the operator learning procedure is \emph{sound}~\cite{konidaris2018skills} over the data, in the following sense: for each transition $\tau = (x, a, x')$, there exists some $\ground{\operator}$, a learned operator ground with objects in $x$, such that $F(\textsc{Abstract}(x, \Psi), \ground{\operator})$ is defined and equals $\textsc{Abstract}(x', \Psi)$. To see this, recall that $\tau$ belongs to an equivalence class, and that this equivalence class was used to learn an operator $\operator$.
Now, we show that the desired $\ground{\operator}$ is $\langle \operator, \delta_\tau \rangle$, where $\delta_\tau$ is the injective parameter-to-object substitution defined above. The $\ground{\controllerspec}$, $\ground{\addeffects}$, and $\ground{\deleteeffects}$ of $\ground{\operator}$ exactly equal those in $\tau$, by construction of the substitution $\delta_\tau$.
Additionally, because $\ground{\preconditions}$ was formed by taking an intersection of abstract states that included $\textsc{Abstract}(x, \Psi)$, it must be the case that $\ground{\preconditions} \subseteq \textsc{Abstract}(x, \Psi)$, since an intersection must be a subset of every constituent set.
By \defref{def:absmodel}, then, the statement is satisfied.
A corollary of this soundness property is that our learned abstractions are guaranteed to obey the semantics we defined in \secref{sec:representation} with respect to the training data.

As a byproduct of operator learning, we have also determined ``local'' datasets for each operator, with each transition in the respective equivalence class defining an example of the operator's preconditions, controller, and effects.
We will use these local datasets and the corresponding substitutions $\delta_\tau$ during sampler learning~\secref{subsec:samplerlearning}.

\subsection{Operator Learning Extended Example}

We conclude our discussion of operator learning with an extended example.
We start with a small toy dataset and use it to walk through each of the three steps in the procedure.

\textbf{Step 1: Generate Dataset.} In this example, our demonstrations contain four transitions, which are tuples $(x, a, x')$. For clarity, we will not write out the task-level states $x$ and $x'$. Additionally, for the sake of the example, we will assume that in this environment there is only one controller \texttt{C}, with no discrete arguments. We abstract these states with the predicate set $\Psi$ includes \texttt{Held}, \texttt{On}, \texttt{IsPurple}, \texttt{IsRed}, \texttt{IsGreen}, \texttt{IsStowable}, and \texttt{IsStowed}, which leads to four $(s, a, s')$ tuples:
\begin{enumerate}
    \item (\{\texttt{On}$(o_1, o_2)$, \texttt{On}$(o_2, o_3)$, \texttt{IsPurple}$(o_{1})$\}, \texttt{C}$(\theta_1)$,\{\texttt{Held}$(o_1)$, \texttt{On}$(o_2, o_3)$,
    \texttt{IsPurple}$(o_{1})$\})
    \item (\{\texttt{On}$(o_4, o_5)$, \texttt{On}$(o_{5}, o_{6})$, \texttt{IsRed}$(o_{4})$\}, \texttt{C}$(\theta_2)$,\{\texttt{Held}$(o_4)$, \texttt{On}$(o_{5}, o_{6})$, \texttt{IsRed}$(o_4)$\})
    \item (\{\texttt{Held}$(o_1)$, \texttt{IsStowable}$(o_1)$,
    \texttt{IsGreen}$(o_2)$\}, \texttt{C}$(\theta_3)$,\{\texttt{IsStowed}$(o_1)$, \texttt{IsStowable}$(o_1)$, \texttt{IsGreen}$(o_2)$\})
    \item (\{\texttt{Held}$(o_8)$, \texttt{IsStowable}$(o_8)$, \texttt{IsGreen}$(o_9)$\}, \texttt{C}$(\theta_4)$,\{\texttt{IsStowed}$(o_8)$, \texttt{IsStowable}$(o_8)$, \texttt{IsGreen}$(o_9)$\})
\end{enumerate}

Intuitively, the first and second transitions might occur when picking up an object ($o_1$ or $o_4$ respectively), while the third and fourth might occur when stowing an object ($o_1$ or $o_8$ respectively). We begin by noting that we can ignore the continuous parameters $\theta_i$ of \texttt{C}, since they do not matter for operator learning (they would be used in sampler learning).

\textbf{Step 2: Produce Equivalence Classes.} Recall that two transitions are in the same equivalence class if there exists a mapping between objects such that the controller, controller discrete arguments, and effects are equivalent. Since we only have one controller \texttt{C} with no discrete arguments in this example, we must only check for effect equivalence. The first transition has effects $(\addeffects, \deleteeffects) = (\{\texttt{Held}(o_1)\}, \{\texttt{On}(o_1,o_2)\})$, while the second has effects $(\addeffects, \deleteeffects) = (\{\texttt{Held}(o_4)\}, \{\texttt{On}(o_4,o_5)\})$. These can be unified with the mapping $\{o_1 \leftrightarrow o_4, o_2 \leftrightarrow o_5\}$. Similarly, the third transition has effects $(\addeffects, \deleteeffects) = (\{\texttt{IsStowed}(o_1)\}, \{\texttt{Held}(o_1)\})$, while the fourth has effects $(\addeffects, \deleteeffects) = (\{\texttt{IsStowed}(o_8)\}, \{\texttt{Held}(o_1)\})$. These can be unified with the mapping $\{o_1 \leftrightarrow o_8\}$.

Note that in this unification procedure, the atoms which were unchanged, such as \texttt{IsPurple}$(o_1)$, do not play a role. Furthermore, the fact that the objects are the same between transitions 1 and 3 is unimportant, because these transitions belong to different equivalence classes.

Selecting an arbitrary transition from each equivalence class and substituting objects with variables, we get the following:
\begin{itemize}
    \item Equivalence class 1:
    \begin{itemize}
    \item $\params$: [\texttt{?x}, \texttt{?y}] \item $\addeffects$: \{\texttt{Held(?x)}\} \item $\deleteeffects$: \{\texttt{On(?x, ?y)}\} \item $\controllerspec$: $\langle$ \texttt{C}, [] $\rangle$
    \item Transitions contained: 1 and 2
    \item $\delta_1$ (substitution for transition 1): \{\texttt{?x} $\to o_1$, \texttt{?y} $\to o_2$\}
    \item $\delta_2$ (substitution for transition 2): \{\texttt{?x} $\to o_4$, \texttt{?y} $\to o_5$\}
    \end{itemize}
    \item Equivalence class 2:
    \begin{itemize}
    \item $\params$: [\texttt{?z}] \item $\addeffects$: \{\texttt{IsStowed(?z)}\} \item $\deleteeffects$: \{\texttt{Held(?z)}\} \item $\controllerspec$: $\langle$ \texttt{C}, [] $\rangle$
    \item Transitions contained: 3 and 4
    \item $\delta_3$ (substitution for transition 3): \{\texttt{?z} $\to o_1$\}
    \item $\delta_4$ (substitution for transition 4): \{\texttt{?z} $\to o_8$\}
    \end{itemize}
\end{itemize}

Note that the parameter list $\params$ for each equivalence class contains all parameters that appear in $\params_{\controllerspec}, \addeffects$, or $\deleteeffects$.

\textbf{Step 3: Learn Operator Preconditions.} We now have all the ingredients of the operators except for their preconditions. For each transition in each equivalence class, we first discard any atom from the abstract state $s$ which involves objects \emph{not} in the image of that transition's substitution $\delta$. For instance, the first transition has $\delta_1 =$ \{\texttt{?x} $\to o_1$, \texttt{?y} $\to o_2$\}. The image is $\{o_1, o_2\}$, which excludes $o_3$. This means that the atom \texttt{On}$(o_2, o_3)$ is discarded from $s$.

After discarding atoms appropriately, we end up with these abstract states for each transition:
\begin{enumerate}
    \item \{\texttt{On}$(o_1, o_2)$, \texttt{IsPurple}$(o_{1})$\}
    \item \{\texttt{On}$(o_4, o_5)$, \texttt{IsRed}$(o_{4})$\}
    \item \{\texttt{Held}$(o_1)$, \texttt{IsStowable}$(o_1)$\}
    \item \{\texttt{Held}$(o_8)$, \texttt{IsStowable}$(o_8)$\}
\end{enumerate}

Now, the preconditions for each equivalence class are obtained by applying each $\delta_i^{-1}$ to these abstract states and taking intersections. This produces the final operator set $\operators$, which does not contain any extraneous atoms related to object color:

\begin{itemize}
    \item Operator 1 (from equivalence class 1):
    \begin{itemize}
    \item $\params$: [\texttt{?x}, \texttt{?y}]
    \item $\preconditions$: \{\texttt{On(?x, ?y)}\}
    \item $\addeffects$: \{\texttt{Held(?x)}\} \item $\deleteeffects$: \{\texttt{On(?x, ?y)}\} \item $\controllerspec$: $\langle$ \texttt{C}, [] $\rangle$
    \end{itemize}
    \item Operator 2 (from equivalence class 2):
    \begin{itemize}
    \item $\params$: [\texttt{?z}]
    \item $\preconditions$: \{\texttt{Held(?z)}, \texttt{IsStowable(?z)}\}
    \item $\addeffects$: \{\texttt{IsStowed(?z)}\} \item $\deleteeffects$: \{\texttt{Held(?z)}\} \item $\controllerspec$: $\langle$ \texttt{C}, [] $\rangle$
    \end{itemize}
\end{itemize}

\subsection{Learning Samplers}
\label{subsec:samplerlearning}

The role of a sampler $\sampler \in \samplers$ is to \emph{refine} its associated operator $\operator$, suggesting continuous parameters of actions that will transition the environment from a state where the preconditions hold to a state where the effects follow.
Recall that a sampler $\sampler : \X \times \O^{|\params|} \to \Delta(\Theta)$ defines a conditional distribution $P(\theta \mid x, o_1, \dots, o_k)$, where $\theta$ are continuous parameters for the controller $C$ in $\operator$, and $(o_1, \ldots, o_k)$ represent a set of objects that could be used to ground $\omega$, with $|\params| = k$.
Using the same demonstration dataset $\D$, we learn samplers of the following form, one per operator: $$\sampler(x, o_1, \dots, o_k) = r_\sampler(x[o_1] \oplus \cdots \oplus x[o_k]),$$
where $x[o]$ denotes the feature vector for $o$ in $x$, the $\oplus$ denotes concatenation, and $r_\sampler$ is the model to be learned.

To learn samplers, we use the local datasets created during operator learning (\secref{subsec:oplearning}), to create datasets for supervised sampler learning, with one dataset per sampler.
Consider any (non-abstract) transition $\tau = (x, a, \cdot)$ in the equivalence class associated with an operator $\omega$.
To create a datapoint for the associated sampler, we can reuse the substitution $\delta_\tau$ found during operator learning to create an input vector $x[\delta_\tau(v_1)] \oplus \cdots \oplus x[\delta_\tau(v_k)]$, where $(v_1, \dots, v_k) = \params$.
The corresponding output for supervised learning is the continuous parameter vector $\theta$ in the action $a$.

With these datasets created, one could use any method for multidimensional distributional regression to learn each $r_\sampler$.
In this work, we learn two neural networks to parameterize each sampler.
The first neural network takes in $x[o_1] \oplus \cdots \oplus x[o_k]$ and regresses to the mean and covariance matrix of a Gaussian distribution over $\theta$; here, we are assuming that the desired distribution has nonzero measure, but the covariances can be arbitrarily small in practice.
This neural network is a sampler in its own right, but its expressive power is limited, e.g., to unimodal distributions.
To improve representational capacity, we learn a second neural network that takes in $x[o_1] \oplus \cdots \oplus x[o_k]$ \emph{and} $\theta$, and returns true or false.
This classifier is then used to rejection sample from the first network.
To create negative examples, we use all transitions $\tau'$ such that the controller in $\tau'$ matches that in $\controllerspec$, but the effects in $\tau'$ are different from $(\addeffects, \deleteeffects)$.

\subsection{Predicate Grammar Details}
\label{app:grammar}

Here we detail the grammar over predicate candidates used in our experiments.
Note that the grammar is the same for all environments (up to the object types and goal predicates).

\begin{tightlist}
\item The base grammar includes two kinds of predicates: all the goal predicates $\Psi_G$, and \emph{single-feature inequality classifiers}. These inequality classifiers are less-than-or-equal-to expressions that compare a constant against an individual feature dimension from $\{1, \ldots, d(\type)\}$, for some object type $\type \in \types$. For the constant, we consider an infinite stream of numbers in the pattern $0.5, 0.25, 0.75, 0.125, 0.375, 0.625, 0.875, \ldots$, which represent \emph{normalized} values of the feature, based on the range of values it takes on across all states in the dataset $\D$. We use this pattern because we want our grammar to describe an infinite stream of classifiers, starting from the median values in $\D$. As an example, a type \texttt{block} might have a feature dimension corresponding to its \texttt{size}, and a classifier could be \texttt{block.size $\leq$ 0.5}. All goal predicates have cost 0. All single-feature inequality classifiers have cost computed based on the normalized constant, with cost 0 for constant $0.5$, cost 1 for constants $0.25$ and $0.75$, cost 2 for constants $0.125$, $0.375$, $0.625$, $0.875$, etc.
\item We include all negations of predicates in the base grammar. Negating a predicate adds a cost of 1.
\item We include two types of universally quantified predicates over the predicates thus far: (1) quantifying over all variables, and (2) quantifying over all but one variable. An example of the first is \texttt{P()} = $\forall$ \texttt{?x}, \texttt{?y} . \texttt{On}(\texttt{?x}, \texttt{?y}), while an example of the second is \texttt{P(?y)} = $\forall$ \texttt{?x} . \texttt{On}(\texttt{?x}, \texttt{?y}). Universally quantifying adds a cost of 1.
\item We include all negations of universally quantified predicates. Negating a predicate adds a cost of 1.
\item Following prior work~\cite{curtis2021discovering}, we prune out candidate predicates if they are equivalent to any previously enumerated predicate, in terms of all groundings that hold in every state in the dataset $\D$. Finally, we discard the goal predicates $\Psi_G$ from the grammar, since they are included in every candidate predicate set  $\Psi$ of our search already.
\end{tightlist}

\subsection{Additional Environment Details}
\label{app:details}

\begin{tightlist}
    \item \textbf{PickPlace1D.} In this toy environment, a robot must pick blocks and place them onto target regions along a table surface. All pick and place poses are in a 1D line. The three object types are block, target, and robot. Blocks and targets have two features for their pose and width, and robots have one feature for the gripper joint state. The block widths are larger than the target widths, and the goal requires each block to be placed so that it completely covers the respective target region, so $\Psi_G = \{\texttt{Covers}\}$, where \texttt{Covers} is an arity-2 predicate. There is only one controller, \texttt{PickPlace}, with no discrete arguments; its $\Theta$ is a single real number denoting the location to perform either a pick or a place, depending on the current state of the robot's gripper. Each action updates the state of at most one block, based on whether any is in a small radius from the continuous parameter $\theta$. Both training tasks and evaluation tasks involve 2 blocks, 2 targets, and 1 robot. In each task, with 75\% probability the robot starts out holding a random block; otherwise, both blocks start out on the table. Evaluation tasks require 1-4 actions to solve. This environment was established by~\citet{silver2021learning}, but that work involved manually defined state abstractions, which we do not provide in this paper.
    \item \textbf{Blocks.} In this environment, a robot in 3D must interact with blocks on a table to assemble them into towers. This is a robotics adaptation of the blocks world domain in AI planning. The two object types are block and robot. Blocks have four features: an x/y/z pose and a bit for whether it is currently grasped. Robots have four features: x/y/z end effector pose and the (symmetric) value of the finger joints. The goals involve assembling towers, so $\Psi_G = \{\texttt{On}, \texttt{OnTable}\}$, where the former has arity 2 and describes one block being on top of another, while the latter has arity 1. There are three controllers: \texttt{Pick}, \texttt{Stack}, and \texttt{PutOnTable}. \texttt{Pick} is parameterized by a robot and a block to pick up. \texttt{Stack} is parameterized by a robot and a block to stack the currently held one onto. \texttt{PutOnTable} is parameterized by a robot and a 2D place pose representing normalized coordinates on the table surface at which to place the currently held block. Training tasks involve 3 or 4 blocks, while evaluation tasks involve 5 or 6 blocks; all tasks have 1 robot. In all tasks, all blocks start off in collision-free poses on the table. Evaluation tasks require 2-20 actions to solve. This environment was established by~\citet{silver2021learning}, but that work involved manually defined state abstractions, which we do not provide in this paper.
    \item \textbf{Painting.} In this challenging environment, a robot in 3D must pick, wash, dry, paint, and place widgets into either a box or a shelf, as specified by the goal. The five object types are widget, box, shelf, box lid, and robot. Widgets have eight features: an x/y/z pose, a dirtiness level (requiring washing), a wetness level (requiring drying), a color, a bit for whether it is currently grasped, and the 1D gripper rotation with which it is grasped if so. Boxes and shelves have one feature for their color. Box lids have one feature for whether or not they are open. Robots have one feature for the gripper joint state. The goals involve painting the widgets to be the same color as either a box or a shelf, and then placing each widget into the appropriate one, so $\Psi_G = \{\texttt{InBox}, \texttt{InShelf}, \texttt{IsBoxColor}, \texttt{IsShelfColor}\}$, all of which have arity 2 (a widget, and either a box or a shelf). There are two physical constraints in this environment: (1) placing into a box can only succeed if the robot is top-grasping a widget, while placing into a shelf can only succeed if the robot is side-grasping it; (2) a box can only be placed into if its respective lid is open. There are six controllers: \texttt{Pick}, \texttt{Wash}, \texttt{Dry}, \texttt{Paint}, \texttt{Place}, and \texttt{OpenLid}. All six are discretely parameterized by a robot argument; \texttt{Pick} is additionally parameterized by a widget to pick up, and \texttt{OpenLid} by a lid to open. \texttt{Pick} has 4 continuous parameters: a 3D grasp pose delta from that widget's center of mass, and a gripper rotation. \texttt{Wash}, \texttt{Dry}, and \texttt{Paint} have 1 continuous parameter each: the amount of washing, the amount of drying, and the desired new color, respectively. \texttt{Place} has 3 continuous parameters: a 3D place pose corresponding to where the currently held widget should be placed. Training tasks involve 2 or 3 widgets, while evaluation tasks involve 3 or 4 blocks; all tasks have 1 box, 1 shelf, and 1 robot. In each task, with 50\% probability the robot starts out holding a random widget; otherwise, all widgets start out on the table. Also, in each task, with 30\% probability the box lid starts out open. Evaluation tasks require 11-25 actions to solve. This environment was established by~\citet{silver2021learning}, but that work involved manually defined state abstractions, which we do not provide in this paper.
    \item \textbf{Tools.} In this challenging environment, a robot operating on a 2D table surface must assemble contraptions by fastening screws, nails, and bolts, using a provided set of screwdrivers, hammers, and wrenches respectively. This environment has physical constraints outside the scope of our predicate grammar, and therefore tests the learner's ability to be robust to an insurmountable lack of downward refinability. The eight object types are contraption, screw, nail, bolt, screwdriver, hammer, wrench, and robot. Contraptions have two features: an x/y pose. Screws, nails, bolts, and the three tools have five features: an x/y pose, a shape, a size, and a bit indicating whether it is held. Robots have one feature for the gripper joint state. The goals involve fastening the screws, nails, and bolts onto target contraptions, so $\Psi_G$ includes \texttt{ScrewPlaced}, \texttt{NailPlaced}, \texttt{BoltPlaced}, \texttt{ScrewFastened}, \texttt{NailFastened}, and \texttt{BoltFastened}. The first three have arity 2 (a screw/nail/bolt and which contraption it is placed on); the last three have arity 1. There are three physical constraints in this environment: (1) a screwdriver can only be used to fasten a screw if its shape is close enough to that of the screw; (2) some screws have a shape that does not match any screwdriver's, and so these screws must be fastened by hand; (3) the three tools cannot be picked up if their sizes are too large. There are eleven controllers: \texttt{Pick\{Screw, Nail, Bolt, Screwdriver, Hammer, Wrench\}}, \texttt{Place}, \texttt{FastenScrewWithScrewdriver}, \texttt{FastenScrewByHand}, \texttt{FastenNailWithHammer}, and \texttt{FastenBoltWithWrench}. All eleven are discretely parameterized by a robot argument; \texttt{Pick} controllers are additionally parameterized by an object to pick up, and \texttt{Fasten} controllers by a screw/nail/bolt and tool (except \texttt{FastenScrewByHand}, which does not have a tool argument). \texttt{Place} has 2 continuous parameters: a 2D place pose corresponding to where the currently held object should be placed, which can be either onto the table or onto a contraption (only if the currently held object is not a tool). Training tasks involve 2 screws/nails/bolts and 2 contraptions, while evaluation tasks involve 2 or 3 screws/nails/bolts and 3 contraptions; all tasks have 3 screwdrivers, 2 hammers, 1 wrench, and 1 robot. Evaluation tasks require 7-20 actions to solve.
\end{tightlist}

\subsection{Additional Experimental Details}
\label{app:additional-experimental-details}
All experiments were conducted on a quad-core Intel Xeon Platinum 8260 processor.
All sampler neural networks are fully connected, with two hidden layers of size 32 each, and trained with the Adam optimizer~\cite{kingma2014adam} for 1K epochs using learning rate 1e-3. The regressor networks are trained to predict a mean and covariance matrix of a multivariate Gaussian; this covariance matrix is restricted to be diagonal and PSD with an exponential linear unit~\cite{clevert2015fast}.
For training the classifier networks, we subsample data to ensure a 1:1 balance between positive and negative examples.
All AI planning heuristics are implemented using Pyperplan~\cite{pyperplan}; all experiments use the LMCut heuristic unless otherwise specified.
The planning parameters are $n_{\text{abstract}} = 1000$ for Tools and $8$ for the other environments, and $n_{\text{samples}} = 1$ for Tools and $10$ for the other environments.

\subsection{Additional Experimental Results}
\label{app:results}

\tabref{tab:learningtimeresults} provides learning times for all experiments.
Tables \ref{tab:allsuccresults}, \ref{tab:allnodesresults}, and \ref{tab:alltimeresults} report success rates, nodes created, and wall-clock time respectively for all evaluation tasks.

\begin{figure*}[b]
  \centering
    \noindent
    \includegraphics[width=\textwidth]{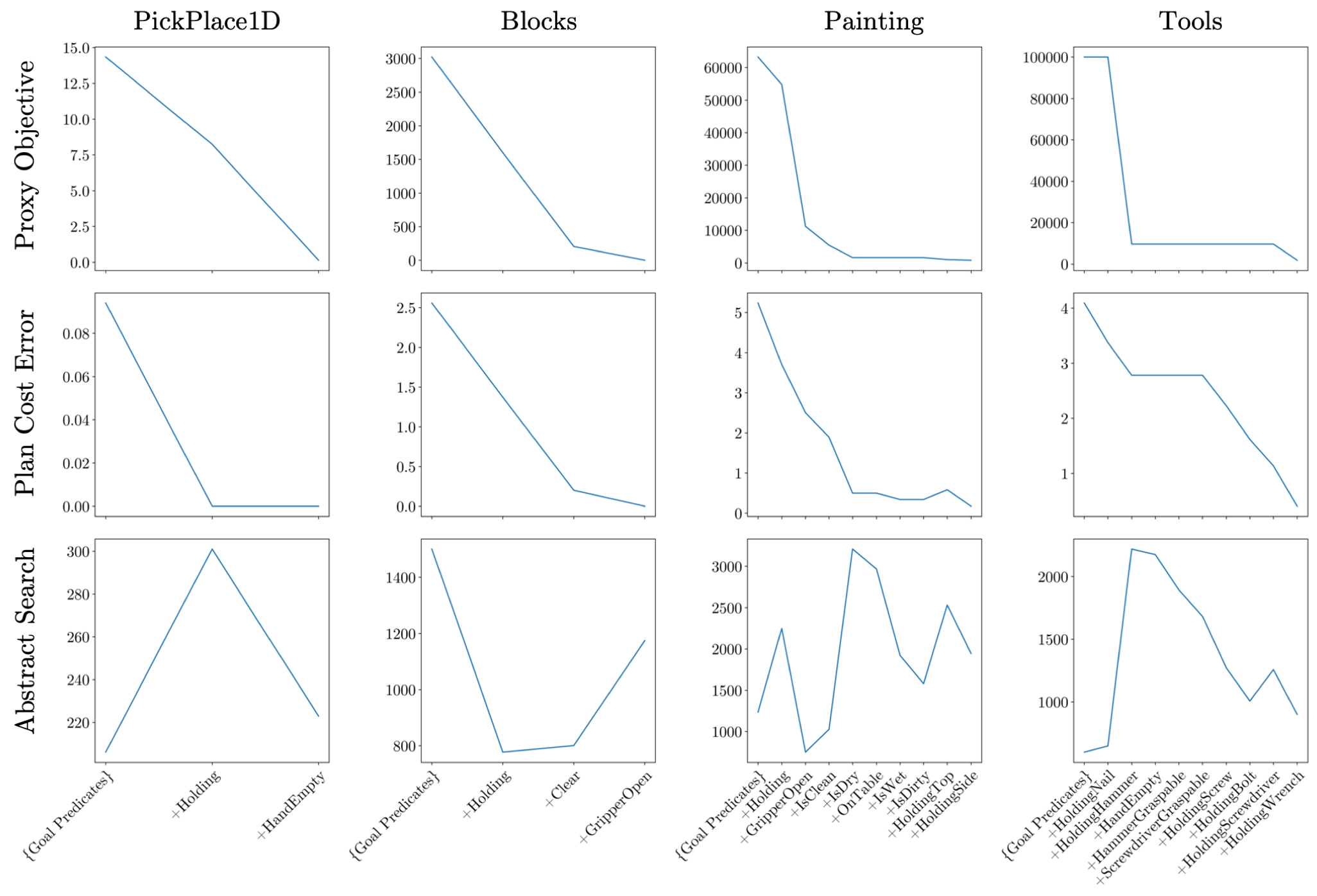}
    \caption{\textbf{Decomposing the surrogate objective.} In these plots, each column corresponds to one environment. The x-axes correspond to sets of manually designed predicates. The predicate sets grow in size from left to right, starting with the goal predicates alone, adding one predicate at each tick mark, and concluding with the full set of manual predicates for the respective environment. The order that the predicates are added was determined by hill climbing with respect to the surrogate objective. The top row shows the surrogate objective itself; the middle row shows the plan cost error $|\textsc{Cost}(\hat{\pi}) - \textsc{Cost}(\pi^*)|$ minimized over the first 8 skeletons generated by abstract search; and the bottom row shows the total number of nodes created by the abstract search (our measure of abstract search time), cumulative over the 8 skeletons. \emph{There are two key takeaways from this plot.} (1) The surrogate objective (first row) monotonically decreases in all environments; this smoothness makes local search over candidate predicate sets an attractive option. (2) Neither of the two components that make up the surrogate objective --- plan cost error (second row) or abstract search time (third row) --- has the same monotonically decreasing property on its own, suggesting that both parts are necessary for making our predicate invention pipeline work. All results are means over 10 seeds.}
  \label{fig:decomposition}
\end{figure*}

\figref{fig:decomposition} analyzes the two main features used by our surrogate objective function. See caption for further description.

We now go through each of our four environments, providing an example of learned predicates and operators from a single seed randomly chosen among successful ones.
We also provide additional statistics for our main method, to supplement the other results we have provided.
Note that the evaluation plan length statistics are averaged over both 10 seeds and 50 evaluation tasks per seed, \emph{with standard deviations over seed only}.

\subsubsection{PickPlace1D}

Statistics for our main method, averaged over 10 random seeds (standard deviations parenthesized):
\begin{tightlist}
\item Average number of predicates in $\Psi$ (both invented and goal predicates): 5.9 (0.54)
\item Average number of operators in $\operators$: 2.1 (0.3)
\item Average plan length during evaluation: 2.44 (0.09)
\end{tightlist}

\begin{figure*}[t]
  \centering
    \noindent
    \includegraphics[width=0.95\textwidth]{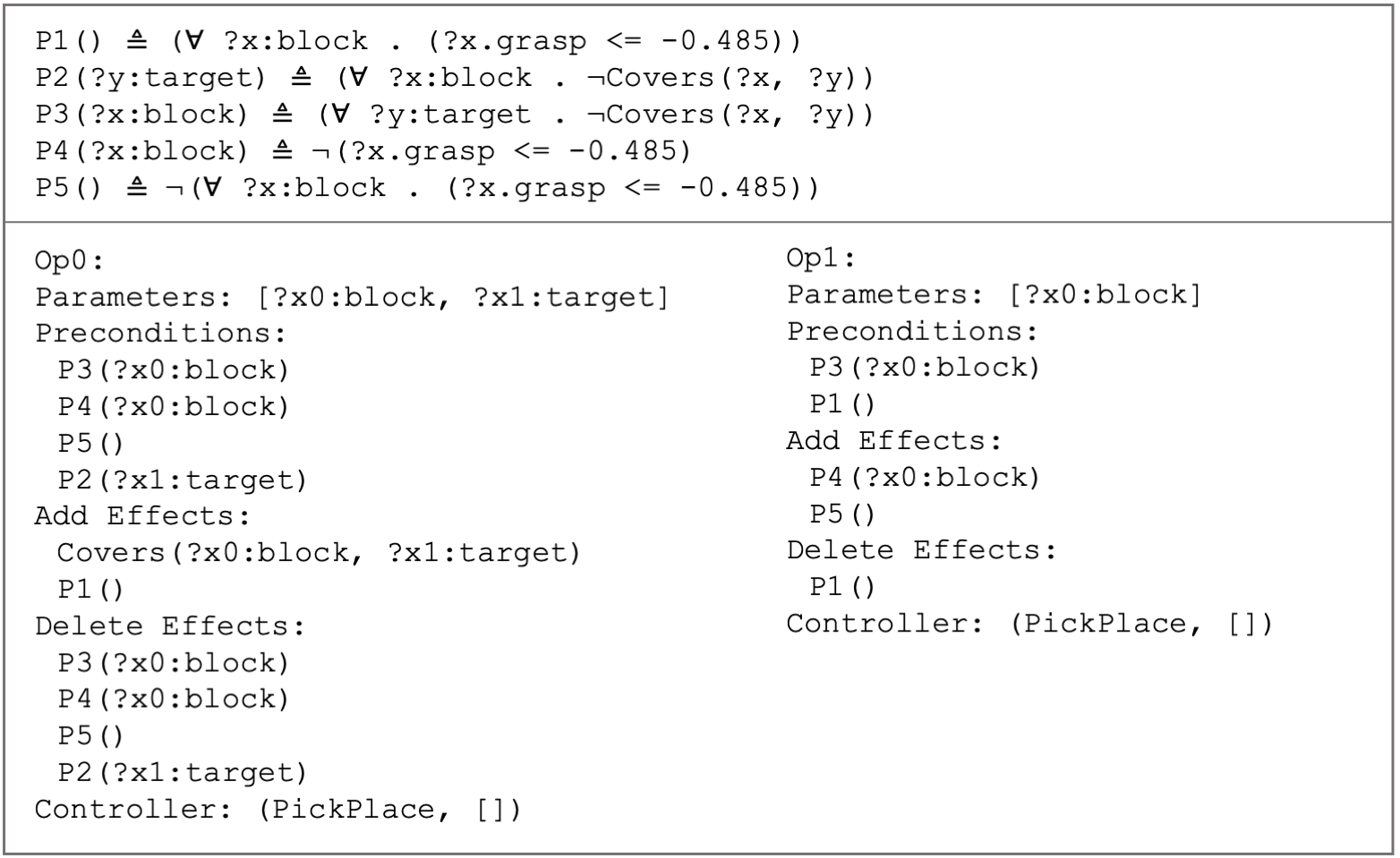}
    \caption{PickPlace1D learned abstractions (top: predicates, bottom: operators).}
  \label{fig:pickplace1dlearned}
\end{figure*}

See \figref{fig:pickplace1dlearned} for example learned predicates and operators for a randomly chosen successful seed.

\subsubsection{Blocks}
Statistics for our main method, averaged over 10 random seeds (standard deviations parenthesized):
\begin{tightlist}
\item Average number of predicates in $\Psi$ (both invented and goal predicates): 6.0 (0.0)
\item Average number of operators in $\operators$: 4.0 (0.0)
\item Average plan length during evaluation: 9.17 (0.69)
\end{tightlist}

\begin{figure*}[t]
  \centering
    \noindent
    \includegraphics[width=0.95\textwidth]{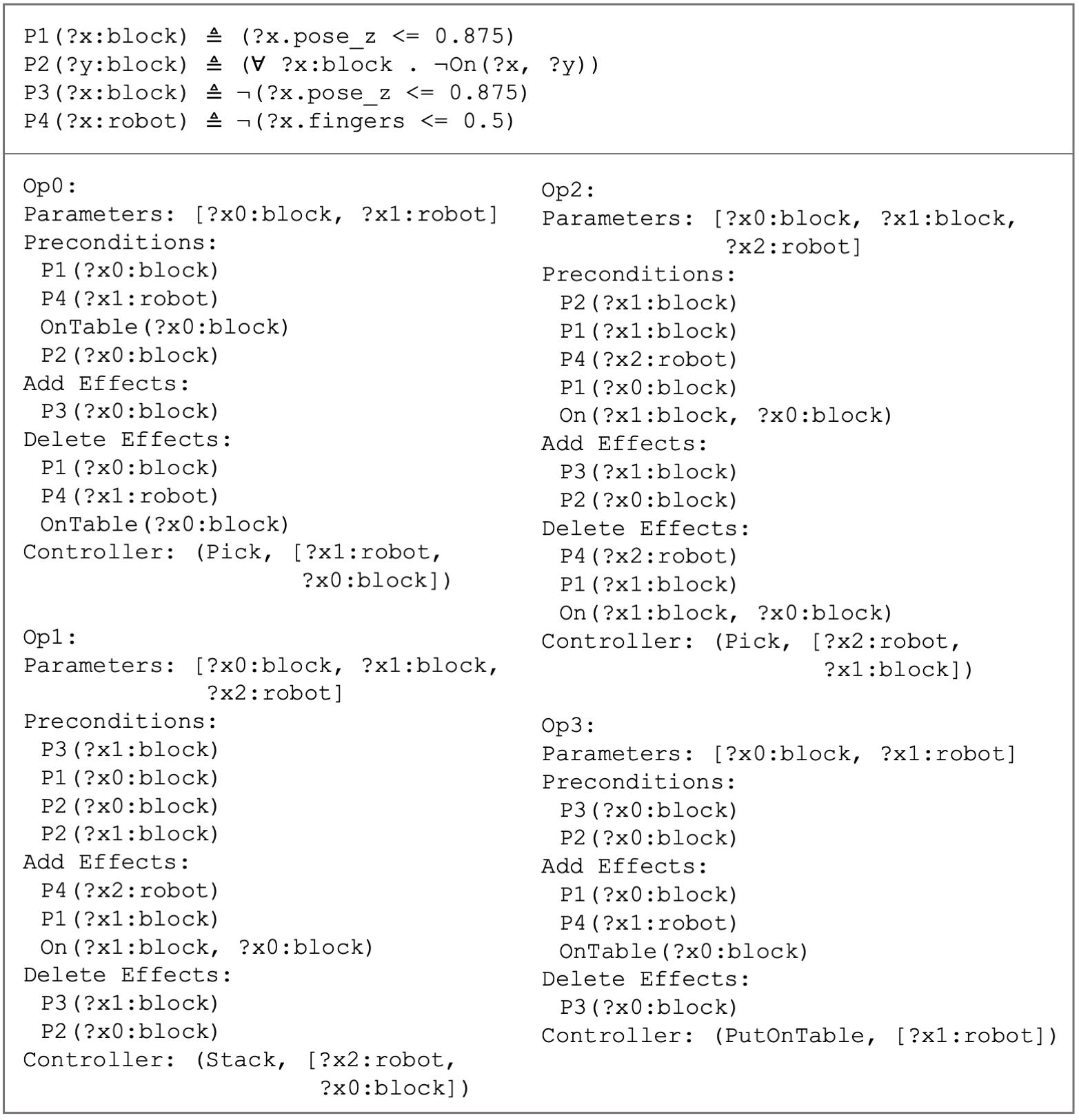}
    \caption{Blocks learned abstractions (top: predicates, bottom: operators).}
  \label{fig:blockslearned}
\end{figure*}

See \figref{fig:blockslearned} for example learned predicates and operators for a randomly chosen successful seed.

\subsubsection{Painting}
Statistics for our main method, averaged over 10 random seeds (standard deviations parenthesized):
\begin{tightlist}
\item Average number of predicates in $\Psi$ (both invented and goal predicates): 22.1 (1.45)
\item Average number of operators in $\operators$: 11.2 (0.6)
\item Average plan length during evaluation: 14.76 (0.29)
\end{tightlist}

\begin{figure*}[t]
  \centering
    \noindent
    \includegraphics[width=0.95\textwidth]{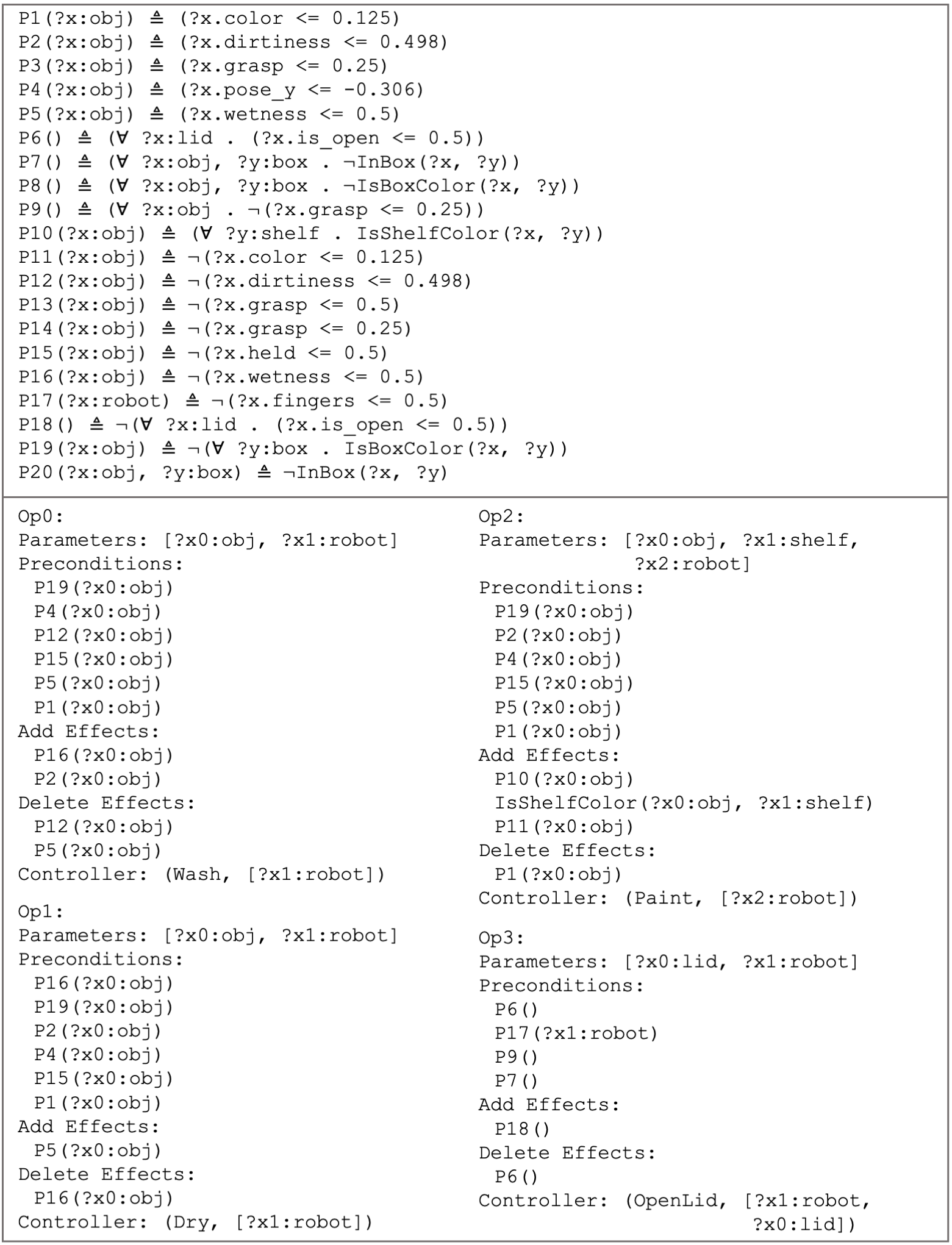}
    \caption{Painting learned abstractions (top: predicates, bottom: operators part 1 of 2).}
  \label{fig:paintinglearned1}
\end{figure*}

\begin{figure*}[t]
  \centering
    \noindent
    \includegraphics[width=0.93\textwidth]{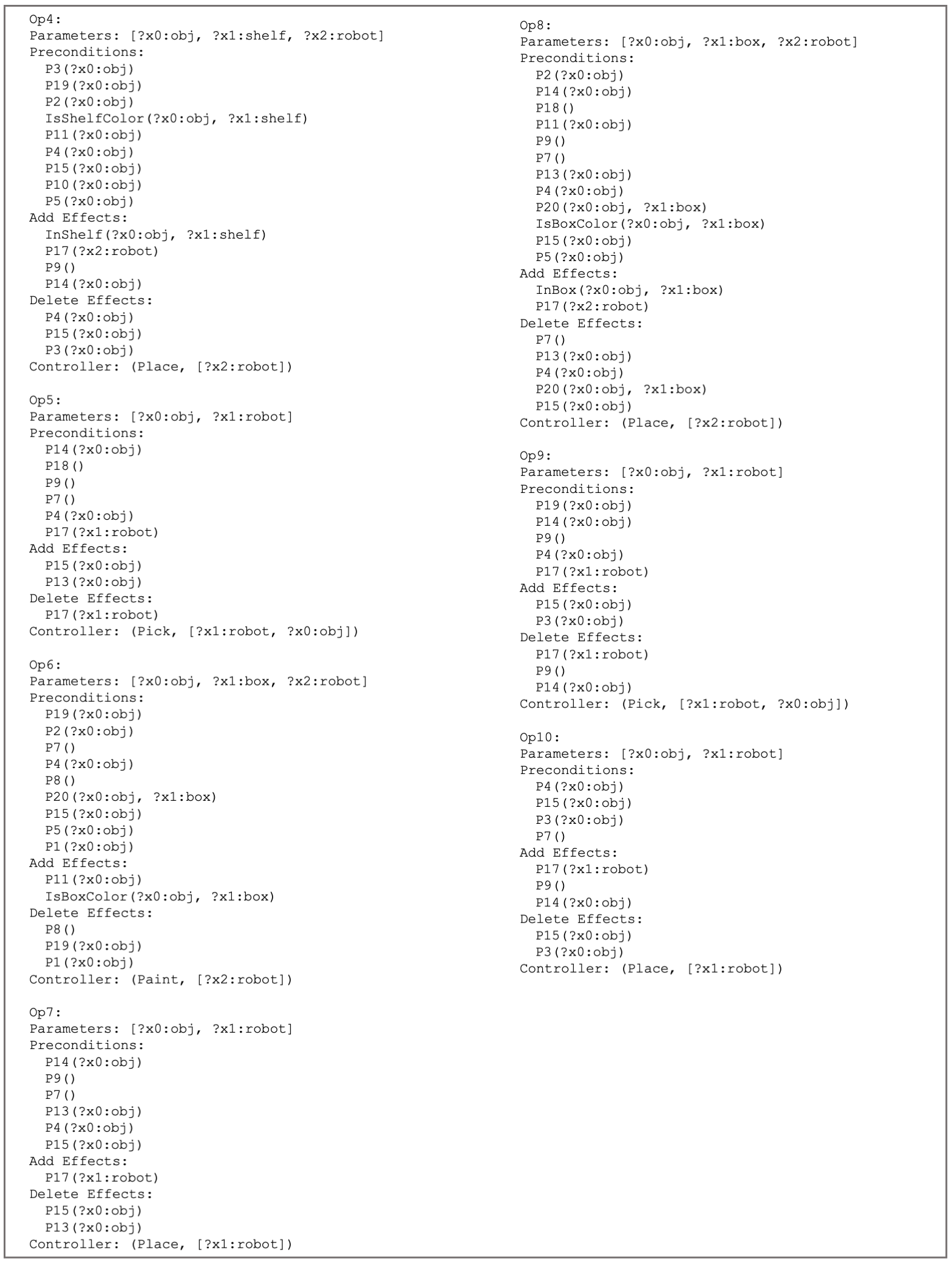}
    \caption{Painting learned abstractions (operators part 2 of 2).}
  \label{fig:paintinglearned2}
\end{figure*}

See Figures \ref{fig:paintinglearned1} and \ref{fig:paintinglearned2} for example learned predicates and operators for a randomly chosen successful seed.

\subsubsection{Tools}
Statistics for our main method, averaged over 10 random seeds (standard deviations parenthesized):
\begin{tightlist}
\item Average number of predicates in $\Psi$ (both invented and goal predicates): 27.4 (4.39)
\item Average number of operators in $\operators$: 17.8 (0.98)
\item Average plan length during evaluation: 10.1 (0.12)
\end{tightlist}

\begin{figure*}[t]
  \centering
    \noindent
    \includegraphics[width=0.95\textwidth]{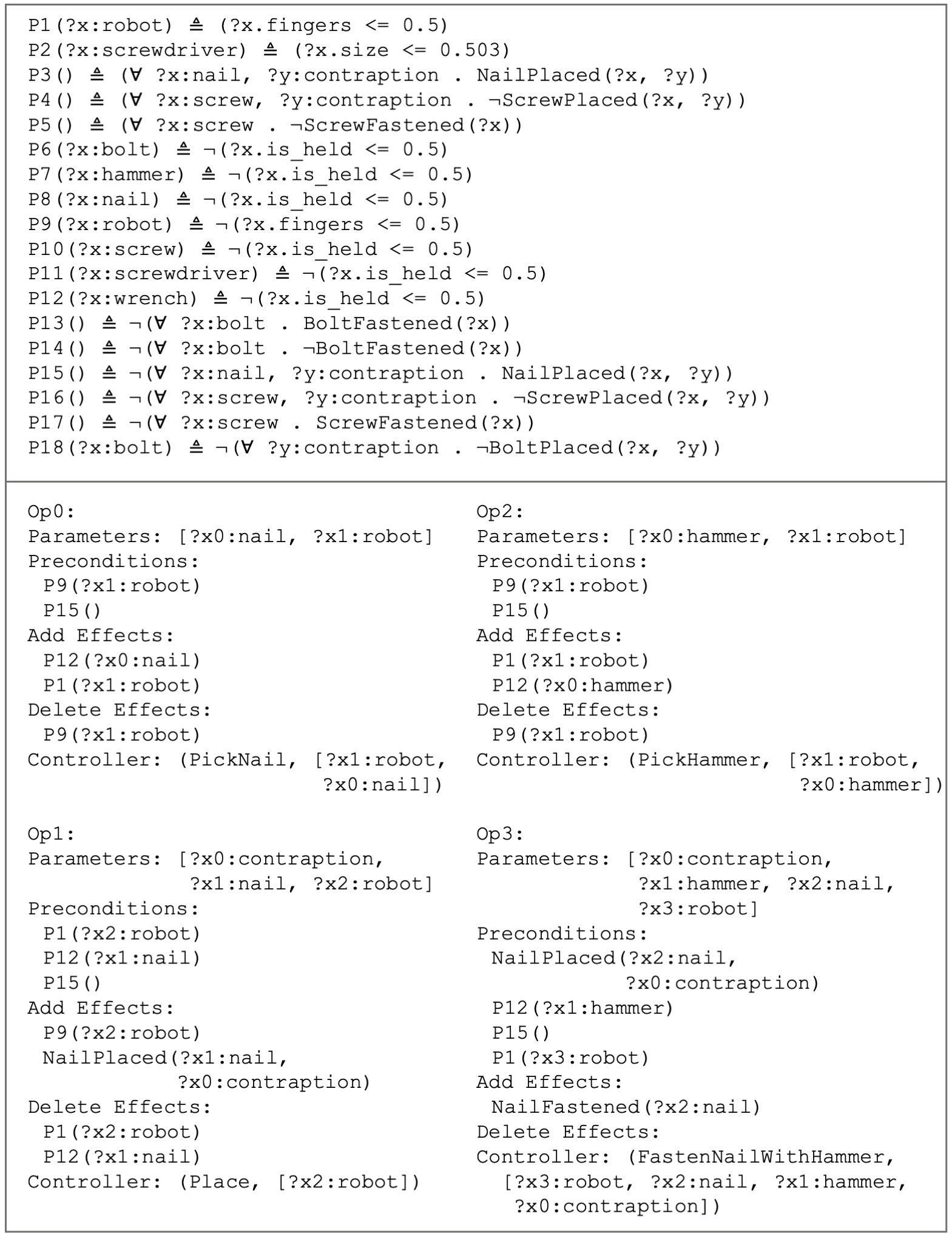}
    \caption{Tools learned abstractions (top: predicates, bottom: operators part 1 of 2).}
  \label{fig:toolslearned1}
\end{figure*}

\begin{figure*}[t]
  \centering
    \noindent
    \includegraphics[width=0.9\textwidth]{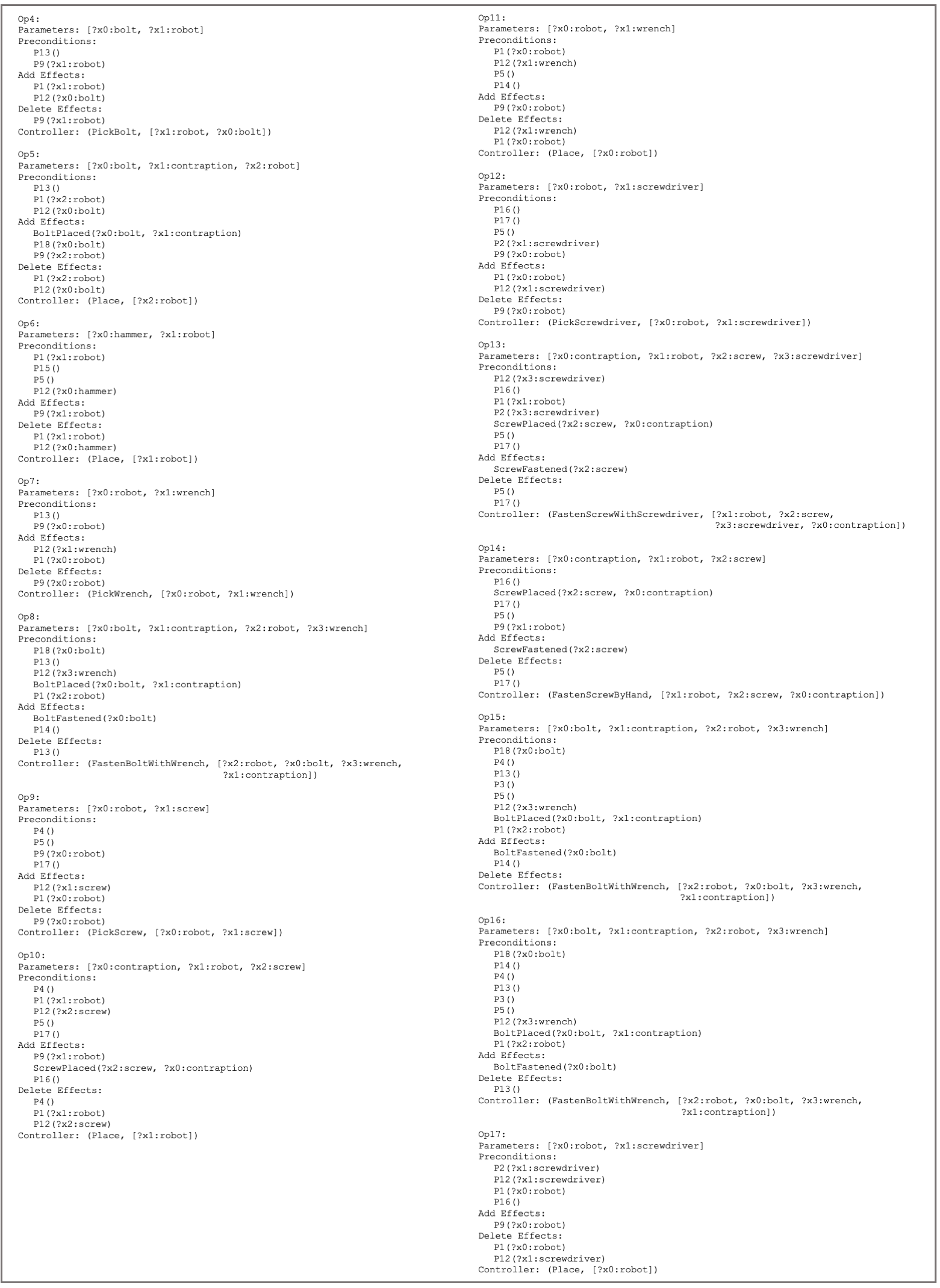}
    \caption{Tools learned abstractions (operators part 2 of 2).}
  \label{fig:toolslearned2}
\end{figure*}

See Figures \ref{fig:toolslearned1} and \ref{fig:toolslearned2} for example learned predicates and operators for a randomly chosen successful seed.

\begin{table}
\begin{center}
\scriptsize
\begin{tabular}{| l | c | c | c | c | c | c | }
	\hline
	\multicolumn{1}{|c|}{} &\multicolumn{3}{c|}{\bf{Ours}} &
	\multicolumn{3}{c|}{\bf{Manual}} \\
	\hline
	\bf{Heuristic} &
	{Succ} & {Node} & {Time}&
	{Succ} & {Node} & {Time}\\
	\hline
	LMCut & 98.4 & 2949 & 0.296 & 98.6 & 2941 & 0.251 \\
	hAdd & 98.6 & 121.6 & 0.115 & 97.8 & 3883 & 0.235 \\\hline
	\end{tabular}
	\caption{\textbf{Varying planning heuristic}. See text for details.}
	\label{tab:varying-heuristic}
\end{center}
\end{table}

\input{learning_time_results_table}
\input{all_succ_results_table}
\input{all_nodes_results_table}
\input{all_time_results_table}

\subsection{Varying the Planner Heuristic}
\label{app:more-blocks-explanations}

\tabref{tab:varying-heuristic} shows an additional experiment we conducted where we varied the AI planning heuristic used by the \textsc{GenAbstractPlan} routine of our bilevel planner in the Blocks environment.
Recall that our predicate invention method uses \textsc{GenAbstractPlan} as well, so it too is affected by this heuristic change. All numbers show a mean over 10 seeds. Interestingly, while the gap in performance is limited when using LMCut, our system shows a massive improvement (over 30x fewer nodes created) versus Manual when using hAdd.
These results are especially surprising because A$^*$ with hAdd is generally considered inferior to other heuristic search algorithms.\footnote{We also experimented with GBFS instead of A$^*$, and hFF, hSA, and hMax instead of hAdd. A$^*$ with hAdd performed best.}
Inspecting the learned abstractions, we find that our approach invents four unary predicates with the intuitive meanings \texttt{Holding}, \texttt{NothingAbove}, \texttt{HandEmpty}, and \texttt{NotOnAnyBlock}, to supplement the given goal predicates \texttt{On} and \texttt{OnTable}.
Comparing these to Manual, which has the same predicates and operators as those in the International Planning Competition (IPC) \cite{ipc2000}, we see the following differences: \texttt{Clear} is omitted\footnote{In the standard encoding, ``clear'' means ``nothing above and not holding.''}, and \texttt{NothingAbove} and \texttt{NotOnAnyBlock} are added.

We observe that \texttt{NothingAbove} and \texttt{NotOnAnyBlock} are logical transformations of predicates used in the standard IPC representation.
This motivated us to run a separate, symbolic-only experiment, where we collected IPC blocks world problems and transformed them to use these learned predicates.
We found that using A$^*$ and hAdd, planning with our learned representations is much faster than planning with the IPC representations.
For example, in the hardest problem packaged with Pyperplan, which contains 17 blocks, planning with our operators requires approximately 30 seconds and 841 node expansions, whereas planning with the standard encoding requires 560 seconds and 17,795 expansions.
We also tried Fast Downward~\cite{helmert2006fast} (again with A$^*$ and hAdd) on a much harder problem from IPC 2000 with 36 blocks.
With our learned representations, planning succeeds in 12.5 seconds after approximately 7,000 expansions, whereas under the standard encoding the planner fails to find a plan within a 2 \emph{hour} timeout.
Note that all of these results are specific to Blocks, A$^*$, and hAdd, and that is exactly the point: even when using an unconventional combination of search algorithm and heuristic, our \emph{planner-aware} method learns abstractions that optimize the efficiency of the given planner in the given environment.

Why exactly do our learned predicates and operators outperform the standard ones when planning with A$^*$ and hAdd? First, we note that it is highly uncommon to use hAdd with A$^*$ in practice with hand-defined PDDL representations, because hAdd is inadmissible and suffers greatly from overestimation issues~\cite{bonet2001planning}. Nevertheless, the interesting phenomenon in our work is that our system is able to \emph{learn an abstraction} that copes with the faults of this combination of search algorithm and heuristic. To understand this further, we make the following observations:
\begin{itemize}
    \item In both cases, the planner must escape from a local minimum with almost every pick operation. For example, in a small problem with 5 blocks where the hand is initially empty, the hAdd values of the states in the plan found are [9, 13, 9, 11, 6, 8, 4, 5, 2, 1, 0] when planning with the standard operators, and [14, 16, 11, 10, 6, 7, 4, 4, 2, 1, 0] when planning with our learned operators. Note the alternation of increasing and decreasing values; the ideal scenario for planning would instead be that these values decrease smoothly.
    \item In states that follow a pick, the hAdd values consistently \emph{overestimate} the true cost-to-go, in both cases. For example, after the first pick with the standard operators, the hAdd value and true cost-to-go are 13 and 9 respectively; for the learned operators, they are 16 and 9 respectively.
    \item Here is the main difference: in states that \emph{precede} a pick, the hAdd values from the standard operators sometimes \emph{underestimate} the true cost-to-go. In the example above, the initial state has an hAdd value of 9, but the true cost-to-go is 10. In harder problems, these underestimations occur with higher frequency; for example, in a problem with 20 blocks, there are 8 cases in the plan found where states preceding picks underestimate the true cost-to-go. In contrast, the hAdd values from our \emph{learned} operators do not ever seem to underestimate the true cost-to-go, in the problems that we analyzed.
    \item Furthermore, this underestimation occurs regularly in states that are local minima, immediately preceding states where the heuristic will be an overestimate, so A$^*$ struggles greatly. Since nodes are expanded in order of $f = g + h$, that is, cost of the plan so far plus heuristic value, A$^*$ will spend time exploring large subtrees rooted at nodes that underestimate true cost-to-go before moving onto the nodes that overestimate it, including those that will ultimately be included in the plan.
\end{itemize}

For reproducibility, we provide the complete operators used to conduct this experiment. We started from the standard blocks domain PDDL downloaded from the \texttt{planning.domains} Github repository, removed the \texttt{Clear} predicate, and added the two predicates our system learned, with the intuitive meanings \texttt{NothingAbove} and \texttt{NotOnAnyBlock}. Problem files were updated accordingly. We ran Fast Downward with the \texttt{--search astar(add)} option. Here is the domain file, with changes highlighted in red (deletion) and green (addition):
\clearpage
\begin{minipage}{0.49\columnwidth}
\begin{Verbatim}[commandchars=\\\{\}]

(define (domain blocks)
(:predicates
    (on ?v0 ?v1)
    (ontable ?v0)
    \textcolor{red}{(clear ?v0)}
    \textcolor{green}{(nothingabove ?v0)}
    \textcolor{green}{(notonanyblock ?v0)}
    (handempty)
    (holding ?v0)
)
(:action pick-up
    :parameters (?x)
    :precondition (and
        \textcolor{red}{(clear ?x)}
        \textcolor{green}{(nothingabove ?x)}
        \textcolor{green}{(notonanyblock ?x)}
        (ontable ?x)
        (handempty))
    :effect (and
        \textcolor{red}{(not (clear ?x))}
        (not (ontable ?x))
        (not (handempty))
        (holding ?x))
)
(:action put-down
    :parameters (?x)
    :precondition (and
        (holding ?x)
        \textcolor{green}{(nothingabove ?x)}
        \textcolor{green}{(notonanyblock ?x)})
    :effect (and
        \textcolor{red}{(clear ?x)}
        (not (holding ?x))
        (handempty)
        (ontable ?x))
)
\end{Verbatim}
\end{minipage}
\begin{minipage}{0.49\columnwidth}
\begin{Verbatim}[commandchars=\\\{\}]
(:action stack
    :parameters (?x ?y)
    :precondition (and
        (holding ?x)
        \textcolor{red}{(clear ?y)}
        \textcolor{green}{(nothingabove ?x)}
        \textcolor{green}{(notonanyblock ?x)}
        \textcolor{green}{(nothingabove ?y)})
    :effect (and
        (not (holding ?x))
        \textcolor{red}{(not (clear ?y))}
        \textcolor{red}{(clear ?x)}
        \textcolor{green}{(not (nothingabove ?y))}
        \textcolor{green}{(not (notonanyblock ?x))}
        (handempty)
        (on ?x ?y))
)
(:action unstack
    :parameters (?x ?y)
    :precondition (and
        (on ?x ?y)
        \textcolor{red}{(clear ?x)}
        \textcolor{green}{(nothingabove ?x)}
        (handempty))
    :effect (and
        (holding ?x)
        \textcolor{red}{(clear ?y)}
        \textcolor{red}{(not (clear ?x))}
        \textcolor{green}{(nothingabove ?y)}
        \textcolor{green}{(notonanyblock ?x)}
        (not (handempty))
        (not (on ?x ?y)))
))
\end{Verbatim}
\end{minipage}

\clearpage

%% file: learning_time_results_table.tex
\begin{table*}[t]
	\centering
	\scriptsize
	
	\begin{tabular}{| l | c || c | c | c | c | c || c | c | }
	\hline
	{\bf{Environment}} &{\bf{Ours}} &
	{\bf{Bisimulation}} &
	{\bf{Branching}} &
	{\bf{Boltzmann}} &
	{\bf{GNN Sh}} &
	{\bf{GNN MF}} &
	{\bf{Manual}} &
	{\bf{No Invent}} \\
	\hline
	PickPlace1D & 625 (134) & 176 (3) & 219 (145) & 264 (17) & 1951 (85) & 1951 (85) & 177 (147) & 66 (0) \\
	Blocks & 10237 (853) & 800 (44) & 1561 (98) & 9798 (1688) & 4047 (209) &4047 (209) & 102 (5) & 84 (2) \\
	Painting & 18395 (28153) & 872 (380) & 2883 (144) & 9457 (3421) & 9185 (166) & 9185 (166) & 565 (457) & 260 (2) \\
	Tools & 18666 (2815) & 573 (20) & 5524 (747) & 9716 (1000) & 7362 (197) & 7362 (197) & 167 (3) & 141 (3) \\\hline
	\end{tabular}
	\caption{\textbf{Learning times in seconds for all experiments}. All numbers are means over 10 seeds, with standard deviations in parentheses. For the GNN-based methods, learning time encompasses training the neural networks. For the other methods, learning time encompasses learning predicates, operators, and samplers (i.e., all components of the abstraction). Even though our main method performs well (Ours), this does come at the cost of increased learning time (although the learning is purely offline). Note that the Manual approach only manually specifies a \emph{state} abstraction (predicates); operators and samplers must still be learned, contributing to the non-zero learning time. Thus, comparing Ours and Manual shows that the large majority of learning time in our system is spent on predicate invention.}
	\label{tab:learningtimeresults}
\end{table*}

%% file: all_succ_results_table.tex
\begin{table*}[t]
	\centering
	\scriptsize
	
	\begin{tabular}{| l | c || c | c | c | c | c | c || c | c | c | }
	\hline
	{\bf{Environment}} &{\bf{Ours}} &
	{\bf{Bisimulation}} &
	{\bf{Branching}} &
	{\bf{Boltzmann}} &
	{\bf{GNN Sh}} &
	{\bf{GNN MF}} &
	{\bf{Random}} &
	{\bf{Manual}} &
	{\bf{Down Eval}} &
	{\bf{No Invent}} \\
	\hline
	PickPlace1D & 98.6 (1.6) & 98.4 (1.5) & 98.4 (1.5) & 98.4 (1.5) & 100.0 (0.0) & 15.2 (8.7) & 19.2 (5.4) & 98.4 (1.5) & 98.6 (1.6) & 39.6 (4.8) \\
	Blocks & 98.4 (1.5) & 19.0 (4.9) & 98.4 (1.5) & 64.8 (23.5) & 27.8 (4.0) & 35.4 (6.8) & 0.6 (0.9) & 98.6 (1.6) & 98.2 (1.4) & 3.2 (2.0) \\
	Painting & 100.0 (0.0) & 0.0 (0.0) & 20.2 (7.1) & 88.6 (29.7) & 59.2 (17.3) & 0.6 (0.9) & 0.0 (0.0) & 99.6 (0.8) & 98.8 (1.8) & 0.0 (0.0) \\
	Tools & 96.8 (4.7) & 26.2 (5.6) & 75.8 (8.4) & 64.2 (3.7) & 25.6 (9.0) & 22.0 (8.9) & 0.0 (0.0) & 100.0 (0.0) & 42.8 (10.4) & 0.0 (0.0) \\\hline

	\end{tabular}
	\caption{\textbf{Percentage of evaluation tasks solved for all experiments}. All numbers are means over 10 seeds, with 50 evaluation tasks per seed, and with standard deviations in parentheses.}
	\label{tab:allsuccresults}
\end{table*}

%% file: all_nodes_results_table.tex
\begin{table*}[t]
	\centering
	\scriptsize
	
	\begin{tabular}{| l | c || c | c | c || c | c | c | }
	\hline
	{\bf{Environment}} &{\bf{Ours}} &
	{\bf{Bisimulation}} &
	{\bf{Branching}} &
	{\bf{Boltzmann}} &
	{\bf{Manual}} &
	{\bf{Down Eval}} &
	{\bf{No Invent}} \\
	\hline
	PickPlace1D & 4.8 (0.2) & 4.7 (0.2) & 4.7 (0.2) & 5.3 (0.2) &    6.5 (0.3) & 4.8 (0.2) & 14.1 (4.0) \\
	Blocks & 2948.5 (1293.2) & 46.9 (18.0) & 2948.5 (1293.2) & 7844.0 (6655.4) &    2940.5 (1299.1) & 2948.5 (1293.2) & 427.7 (83.7) \\
	Painting & 501.8 (180.0) & \;\;\;-- & 876.6 (509.7) & 4008.8 (3851.3) &    2607.5 (1117.2) & 489.0 (190.2) & \;\;\;-- \\
	Tools & 1897.2 (1404.0) & 5247.7 (2560.6) & 167.8 (78.4) & 909.9 (174.1) &    4770.9 (886.8) & 152.5 (27.6) & \;\;\;-- \\\hline
	\end{tabular}
	\caption{\textbf{Number of nodes created by abstract search during planning in evaluation tasks}. All numbers are means over \emph{solved tasks only} across 10 seeds, with 50 evaluation tasks per seed, and with standard deviations in parentheses.}
	\label{tab:allnodesresults}
\end{table*}

%% file: all_time_results_table.tex
\begin{table*}[t]
	\centering
	\scriptsize
	
	\begin{tabular}{| l | c || c | c | c | c | c | c || c | c | c | }
	\hline
	{\bf{Environment}} &{\bf{Ours}} &
	{\bf{Bisimulation}} &
	{\bf{Branching}} &
	{\bf{Boltzmann}} &
	{\bf{GNN Sh}} &
	{\bf{GNN MF}} &
	{\bf{Random}} &
	{\bf{Manual}} &
	{\bf{Down Eval}} &
	{\bf{No Invent}} \\
	\hline
	PickPlace1D & 0.006 (0.0) & 0.006 (0.0) & 0.006 (0.0) & 0.005 (0.0) & 0.436 (0.1) & 0.014 (0.0) & 0.004 (0.0) & 0.045 (0.0) & 0.008 (0.0) & 1.369 (0.6) \\
	Blocks & 0.296 (0.1) & 0.158 (0.1) & 0.284 (0.1) & 0.954 (0.3) & 0.138 (0.1) & 0.249 (0.1) & 0.006 (0.0) & 0.251 (0.1) & 0.318 (0.1) & 1.235 (1.3) \\
	Painting & 0.470 (0.2) & \;\;\;-- & 4.186 (0.9) & 0.600 (0.3) & 2.077 (1.2) & 0.073 (0.0) & \;\;\;-- & 0.464 (0.1) & 0.208 (0.0) & \;\;\;-- \\
	Tools & 0.457 (0.3) & 0.699 (0.3) & 0.109 (0.0) & 0.247 (0.0) & 0.311 (0.2) & 0.043 (0.0) & \;\;\;-- & 0.491 (0.1) & 0.060 (0.0) & \;\;\;-- \\\hline

	\end{tabular}
	\caption{\textbf{Total time in seconds for evaluation tasks}. These results encompass planning time (when applicable) and policy or plan inference time (the time taken to produce an action at each step, given the current state). All numbers are means over \emph{solved tasks only} across 10 seeds, with 50 evaluation tasks per seed, and with standard deviations in parentheses.}
	\label{tab:alltimeresults}
\end{table*}